\pdfoutput=1
\documentclass[11pt]{article}

\usepackage{acl}

\usepackage[T1]{fontenc}
\usepackage[utf8]{inputenc}

\usepackage{microtype}

\usepackage{inconsolata}
\usepackage{times}
\usepackage{array}
\usepackage{amsmath}
\usepackage{siunitx}
\usepackage{latexsym}
\usepackage{booktabs}
\usepackage{longtable}
\usepackage{multirow}
\usepackage{graphicx}
\usepackage{graphics}
\usepackage{caption}
\usepackage{subcaption}
\usepackage{xcolor}
\usepackage{multicol}
\usepackage{colortbl}
\usepackage{url}
\usepackage{xurl}
\usepackage{tabularx}
\usepackage{algorithm}
\usepackage{algpseudocode}
\usepackage{soul}
\usepackage{tcolorbox}

\definecolor{top_performing}{RGB}{200, 247, 186}
\definecolor{runner_up}{RGB}{193, 236, 250}
\definecolor{dark_green}{RGB}{200, 247, 186}
\title{\textsc{Standardize}: Aligning Language Models with Expert-Defined Standards for Content Generation}

\author{Joseph Marvin Imperial$^{\Omega,\Lambda}$~\;~Gail Forey$^{\Lambda}$~\;~Harish Tayyar Madabushi$^{\Lambda}$ 
\\$^{\Lambda}$University of Bath, UK\\ $^{\Omega}$National University, Philippines
\\\texttt{\href{mailto:jmri20@bath.ac.uk}{jmri20@bath.ac.uk}~\;~\href{mailto:gf370@bath.ac.uk}{gf370@bath.ac.uk}~\;~\href{mailto:htm43@bath.ac.uk}{htm43@bath.ac.uk}}
}

\begin{document}
\maketitle
\begin{abstract}

Domain experts across engineering, healthcare, and education follow strict standards for producing quality content such as technical manuals, medication instructions, and children's reading materials. However, current works in controllable text generation have yet to explore using these standards as references for control. Towards this end, we introduce \textsc{Standardize}, a retrieval-style in-context learning-based framework to guide large language models to align with expert-defined standards. Focusing on English language standards in the education domain as a use case, we consider the Common European Framework of Reference for Languages (CEFR) and Common Core Standards (CCS) for the task of open-ended content generation. Our findings show that models can gain $45$\% to $100$\% increase in precise accuracy across open and commercial LLMs evaluated, demonstrating that the use of knowledge artifacts extracted from standards and integrating them in the generation process can effectively guide models to produce better standard-aligned content.\footnote{Code and data: \texttt{\url{https://github.com/imperialite/standardize-ctg}}}
\end{abstract}

%%
%% - FIGURE 1 - SPECGEM PROPOSAL
%%
\begin{figure}[!t]
    \centering
    \includegraphics[width=.49\textwidth]{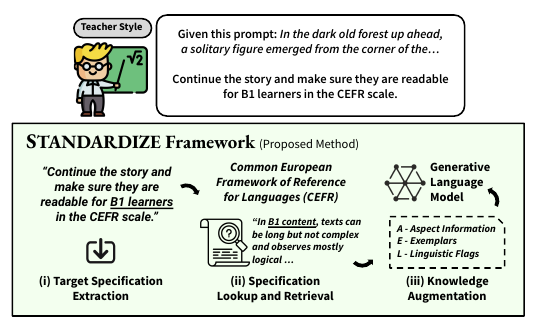}
    \caption{In contrast to the simple prompting method used by teachers, the proposed \textsc{Standardize} framework aims to improve the performance of generative models for content generation by using the fine-grained information found in expert-defined standards. The framework involves a three-part process starting with the (i) \textbf{extraction} of target specifications from the prompt, (ii) \textbf{lookup and retrieval} of information that matches the target specifications from the specified standard, and (iii) \textbf{knowledge augmentation} to produce artifacts that represent the standard itself for integration into the generation process with generative models.}
    \label{fig:fig1}
\end{figure}

\section{Introduction}
One of the most realized benefits of large language model (LLM) research is how it became widely adopted by the public. In particular, the rise of chat-style model interfaces, such as ChatGPT and Perplexity, has allowed non-technical users to fully utilize these tools in accomplishing day-to-day tasks and activities, such as getting help with writing, documenting code, and providing recommendations. A key technological advancement behind this is the use of reward-based methods such as Reinforcement Learning for Human Feedback (RLHF, \citet{ouyang2022training}), which embeds human preferences to generative models for better-aligned outputs with respect to the task at hand. 
%{while RLHF is used for alignment, standards are not used by this method. Link this gap.}

Despite the growing literature of complex algorithms and architectures for enriching the instruction-following capabilities of LLMs, the missing puzzle piece that seems to have not garnered equal attention from the community is the integration of actual {standards} or {guidelines} crafted by domain experts as a reference of control. For example, in education and language assessment, standards such as the Common European Framework of Reference for Languages (CEFR) serve as an accredited guide for administrators in charge of the creation of educational curriculum content. This standard provides fine-grained specifications of text complexity that different levels of learners can understand depending on their language proficiency \cite{north2007cefr,north2014cefr}. To be able to automatically generate text content (e.g., narratives or short stories) using an LLM that is acceptable by CEFR standards and captures a student's topic interest at the same time can serve as a powerful tool in classroom engagement for educators in the long run. Thus, this research gap is an opportunity where the complex instruction-following capabilities of language models can provide assistance, particularly for tasks requiring the generation of text content since this is one of the areas where these models objectively perform well \cite{chung2022scaling,wei2021finetuned,gatt2018survey}.

%in healthcare and engineering, well-documented standards are strictly followed in order to ensure the quality of processes. This includes the UK National Health Service (NHS) Injectable Medicines Guide (IMG) which contains instructions on how medical injectables should be mixed \cite{keeling2010injectable} as well as the Simplified Technical English (STE)\footnote{\url{https://www.asd-ste100.org/}} which is a documented controlled language specification for writing technical manuals that are simple to read. Following these standards, even for domain experts, can be tedious, challenging, and even consequential in serious cases due to its complexity \cite{jones2021user,cousins2005medication}. 

Towards this end, we tackle the main research question: \textbf{How can we align large language models for content generation tasks using expert-defined standards?} We list our major contributions from this study as follows:

\begin{enumerate}
    \item We introduce \textsc{Standard-CTG}, a new task formalizing the challenge of generating text using generative language models with expert-defined standards as a for controllability.

    \item We propose \textsc{Standardize}, a new retrieval-style in-context learning framework that extracts knowledge artifacts from standards such as aspect information, exemplars, and manually crafted linguistic variables to improve the performances of generative language models for content generation.

    \item We introduce significantly improved performances for GPT-4 and Llama for the task of \textsc{Standard-CTG} using two of the most widely recognized academic standards, CEFR and CCS, across diverse evaluation procedures.
\end{enumerate}

\section{Expert-Defined Standards}

\subsection{Background}
According to the International Organization for Standardization (ISO)\footnote{\url{https://www.iso.org/standards.html}}, \textbf{standards} are documented guidelines often containing rich detail in describing requirements, specifications, and criteria. These guidelines are defined and continuously improved by experts in various domains, such as education, healthcare, and accounting, to name a few. Using standards ensures an institution's products and processes are consistent and reproducible \cite{sadler2017academic}. 

In the context of education and language assessment, standards are usually in the form of either (a) {content standards} such as documentations of a common language for ease of communication, writing, and content production, and (b) {performance standards} such as state-administered tests for reading and mathematical problem-solving competencies. This study focuses on content-based standards used in education and language assessment to be integrated into a generative model's text generation process. The alignment with existing standards for any generated text material is crucial to ensure quality and consistency before being used in classroom settings \cite{la2000state}.

\subsection{Standards in Education and Language Assessment}
We discuss the two selected English standards we consider as test cases for this study.\\
 
\noindent \textbf{The Common European Framework of Reference for Languages (CEFR)} is one of the well-known standard language framework\footnote{\url{https://www.coe.int/en/web/common-european-framework-reference-languages/level-descriptions}} developed by The Council of Europe and used for assessing general language competencies such as reading, writing, and listening \cite{north2007cefr,north2014cefr}. The CEFR uses a six-point level scale of A1, A2, B1, B2, C1, and C2, which denotes increasing complexities in instructional content development. We use the level descriptors compiled by \citet{natova2021estimating}, which cover three aspects, namely (1) Meaning/Purpose, (2) Structure, and (3) Grammatical Complexity, describing the characteristics of desired content per level as shown in Table~\ref{tab:full_main_specs_table}. We omit a fourth aspect of Reader's Knowledge Demands from the standard as this heavily depends on the reader's background knowledge and is entirely subjective \cite{forey2020whole,forey2019benefits}. \\

\noindent\textbf{The Common Core Standards (CCS)} is an academic standard\footnote{\url{https://corestandards.org/}} developed by the US National Governors Association and the Council of Chief State School Officers (CCSSO) which has been widely adopted by schools across the United States for its K-$12$ curriculum. In this study, we adapt the recommended model of CCS for assessing text complexity, which includes two main variables: (1) Qualitative Dimensions and (2) Quantitative Dimensions. However, similar to the CEFR standard, we do not include the last variable, which is Reader Considerations, as this requires professional judgment or a teacher's intervention. The description of each aspect of CCS is detailed in Table~\ref{tab:full_main_specs_table}.

\section{Standard-Aligned Content Generation (\textsc{Standard-CTG})}
\label{sec:standard-ctg}

Given the importance of adhering to expert-defined standards in the context of language assessment, we introduce a new task we refer to as \textit{\bf standard-aligned content generation} (\textsc{Standard-CTG}). The overarching goal of \textsc{Standard-CTG} is to pave the way for new approaches that aim to integrate the conventional methodologies of controllable text generation in NLP with actual constraints provided by domain experts across interdisciplinary fields such as education, engineering, and medicine through documented standards. To align with terminologies used in education and other non-computing literature, in this work, we use the term \textit{content generation} instead of \textit{text generation} as usually seen in technical NLP literature. 

We represent the task of \textsc{Standard-CTG} using the following formulation:

%\begin{equation}
%    \textsc{A} = C_{\mathrm{Stndrd}}(M(p,a,k_a),E)
%\end{equation}

%where $\textsc{A}$ quantifies the content alignment score of using a general evaluator $C_{\mathrm{Stndrd}}$ that tests the quality of a language model's $M$ generated content against a collection of gold-standard examples $E$ using inputs such as (a) a natural language prompt $p$, (b) information of some aspect $a$, and (c) transformed representation of an aspect $k_a$ defined or extracted from the chosen standard. We pattern our major experiments in the succeeding sections based on this formulation.
\begin{equation}
\begin{split}
    \text{\textsc{Standard-CTG}}(\mathbf{X},\mathbf{D}_\mathrm{Standard}) \\ = \mathcal{L}(\mathcal{M}_{\theta}(\mathbf{X}, \mathbf{\tilde{K}}_\mathrm{Standard}),\mathbf{E})
\end{split}
\end{equation}

where $\mathcal{L}$ is a general evaluator that tests how close a language model's $\mathcal{M_\theta}$ generated content $\mathbf{X}$ is with gold-standard examples $\mathbf{E}$ through learning transformed knowledge representations  $\mathbf{\tilde{K}}_\mathrm{Standard}$ of the selected standard $\mathbf{D}_\mathrm{Standard}$. The evaluator $\mathcal{L}$ can assume many forms, including model-based, distance-based, and reference-based scoring. We pattern our major experiments in the succeeding sections based on this formulation.

%We define this task as a derivation of the bigger research area of controllable text generation, but with the addition of an external reference being the chosen standard to consider. 

\section{The \textsc{Standardize} Framework}
\label{sec:standardize_framework}

Given that expert-defined standards are naturally information-rich, lengthy, and complex, our main hypothesis in this study is that in order for a generative language model to produce content that is \textit{aligned} with the specifications provided by a standard, the information found in the standard must be considered in the generation process. The challenge then is redirected towards \textit{how} any information extracted can be represented as something that the generative model will find useful. 

Towards addressing \textsc{Standard-CTG}, we propose \textsc{Standardize}, a retrieval-style in-context learning-based framework that exploits the rich information found in standards and transforms this into knowledge artifacts to improve the quality of content produced by generative models. Figure~\ref{fig:fig1} encapsulates this framework in a visual manner. In the succeeding sections, we discuss the proposed \textsc{Standardize} framework more thoroughly.  \\

%\subsection{Process}
%\label{sec:process}

\noindent \textbf{Target Specification Extraction} is performed first to obtain informative tags in the prompt and to correctly match this information within the standards. For academic standards in language assessment, these specifications should provide information about  \textit{who} will be content delivered to (target audience) and using \textit{what} specific standard out of many (CEFR or CCS). Thus, these two information tags are the basic required input for the process. As an example shown in Figure~\ref{fig:fig1}, the extracted specifications provided in the prompt are \textit{A2 readers}, which points to a particular group of learners requiring low-leveled reading materials, and \textit{CEFR scale}, which denotes the selected standard where properties of A2-level texts are described. \\

\noindent \textbf{Specification Lookup and Retrieval} is then performed next upon extracting the target specifications. A lookup process is done to find a match with the selected standard, usually in the form of a database or an external machine-readable file. In this work, we simply transformed the level-specific descriptors from \citet{natova2021estimating} into a .csv file. The information from the standard in the form of \textit{aspects} (or characteristics) that match the target specifications is then retrieved. The length and complexity of a standard's level of information regarding its specifications may vary. As shown in Figure~\ref{fig:fig1} for the CEFR standard, the retrieved information that matches the desired level of complexity for the target audience (A2 readers) can be checked at Table~\ref{tab:full_main_specs_table}. \\

%In the conventional method of prompting generative models to produce content, supplying surface-level detail is the easiest method and is particularly favored by non-technical users. Figure~\ref{fig:fig1} shows the simple prompting method for generative model interfaces such as ChatGPT, which users such as teachers recommend \cite{Staake_2023}. 

\noindent \textbf{Knowledge Augmentation} is done last but is the most important process of the pipeline. We propose a further technical augmentation of information found in standards to obtain \textbf{knowledge artifacts} in the prompts. These knowledge artifacts can range from simple additional information already present in the standard to complex representations, such as incorporating actual linguistic features to control the granularity of the generation process. Recent works surveying the performance of open and closed models have shown that non-informative style of prompting language models, such as the teacher style shown in Figure~\ref{fig:fig1}, is effective only to a certain extent and may be biased towards content generation in lower levels, such as A2 or B1 in the CEFR standards \cite{imperial-madabushi-2023-uniform,ribeiro-etal-2023-generating}. 

\section{Knowledge Artifacts for \textsc{Standardize}}
\label{sec:knowledge_artifacts}
In this section, we discuss the knowledge artifacts $\mathbf{\tilde{K}}_\mathrm{Standard}$ extracted from the two educational standards  $\mathbf{D}_\mathrm{Standard}$ used in the \textsc{Standardize} framework and how they are integrated into the generation setup via in-context learning. \\
%While this study specifically makes use of CEFR and CCS, which are academic standards for language assessment and content development, these knowledge artifacts may also be present across standards in interdisciplinary fields. For example, in aerospace industries engineering, a standard called ASD-STE100 Simplified Technical English (STE)\footnote{\url{https://www.asd-ste100.org/}} is used to document and create technical manuals using a controlled language. Thus, we note that the following artifacts may be scaled and applied to other standards for aligned content generation using generative models.\\

\noindent \textbf{{Baseline (Teacher Style)}} We treat the Teacher Style method as seen in Figure~\ref{fig:fig1}, where a simple, non-enriched prompt contains the target category from each standard, as the baseline for performance. We use this term in observance of how non-technical users, especially teachers, interact with generative chat interfaces \cite{imperial2023flesch}. \\

\begin{figure}[!t]
    \centering
    \includegraphics[width=.45\textwidth,trim={2cm 5cm 8cm 2cm}, clip]{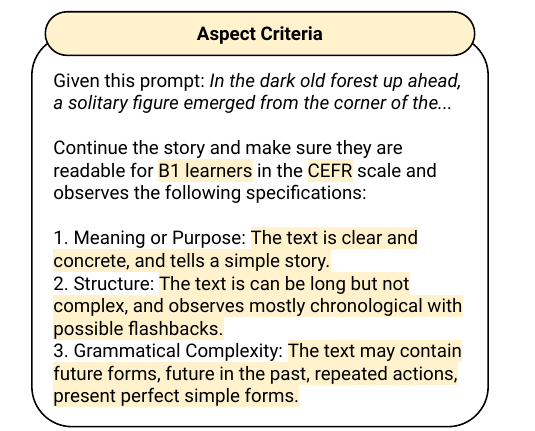}
    \caption{A standard contains recommended characteristics of content across one or more \textbf{domain-specific aspects} or criteria. This figure shows an example of the CEFR standard where the set of criteria includes depth of meaning, structure, and grammatical complexity.}
    \label{fig:aspect_criteria}
\end{figure}

\noindent \textbf{Aspect Information (\textbf{\textsc{Standardize}}-$\boldsymbol{A}$)} represents the specific descriptive information provided in the standard. In the context of standards for content generation, aspect information is generally attributed to linguistic criteria of content with respect to its target audience. Figure~\ref{fig:aspect_criteria} shows how aspect information from a standard (e.g., CEFR) can be integrated into the actual prompt. The addition of aspect criteria information ensures that the generative model will have access to \textit{explicit characteristics} of the desired generated content in different dimensions. \\

%%
%% - ASPECT CRITERIA
%%

%%
%% - LINGUISTIC FLAGS
%%
\begin{figure}[!t]
    \centering
    \includegraphics[width=.45\textwidth,trim={2cm 4cm 8cm 2cm}, clip]{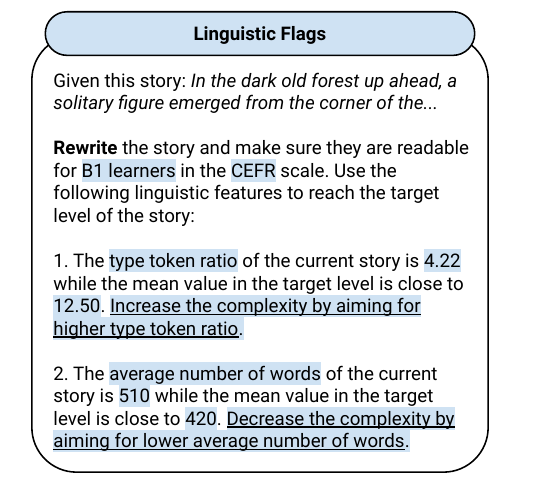}
    \caption{A standard contains aspect definition which can be represented by flags such as \textbf{linguistic variables}. Given the mean values from gold-standard data in the target level, the generative model can then be steered to push the property of its generated content using \textbf{directional instructions} such as \textit{increase} or \textit{decrease}.}
    \label{fig:linguistic_flags}
\end{figure}

%%
%% - EXEMPLARS
%%
\begin{figure}[!t]
    \centering
    \includegraphics[width=.43\textwidth,trim={2cm 6.8cm 8cm 2cm}, clip]{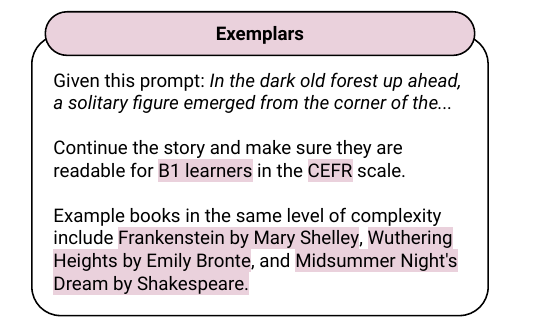}
    \caption{A standard contains recommended \textbf{exemplars} that serve as gold-standard reference. This figure shows an example of the CEFR standard where three well-known pieces of literature are provided as examples of content that conforms to the target level specified (B1).}
    \label{fig:exemplars}
\end{figure}

\noindent \textbf{Linguistic Flags (\textbf{\textsc{Standardize}}-$\boldsymbol{L}$)} represent the controllable attribute-based variables of a standard that a generative model can use to steer the direction of content generation. In the \textsc{Standardize} framework, this process serves as a \textbf{rewrite function} where a generative model is asked to produce an initial content first using another method prompting (e.g., aspect information in Figure~\ref{fig:aspect_criteria}), and rewrites this by comparing linguistic flag values of the initially generated content against the mean value of a gold standard dataset of the target level. An example is illustrated in Figure~\ref{fig:linguistic_flags} where the mean type-token ratio of a collection of gold-standard B1-level text $12.5$ is added to the prompt while being compared to the current type-token value of the story, which is $4.2$. A \textit{verbalizer} is used to transform the computed linguistic flags into natural language prompts. The keywords \textit{increase} and \textit{decrease} are used in constructing the prompts to provide a sense of direction for the generative model. 

In this work, we select $2$ to $4$ linguistic flags for both CEFR and CCS as reported in Table~\ref{tab:full_main_specs_table}. The selection of what linguistic flags to use can be as simple as referring to what the definitions of aspects provide and need not be exhaustively many. For example, in CEFR, the Organization aspect is defined through different levels as \textit{"text is often short and observes chronological and predictable structure"} for A2 and \textit{"text is can be long but not complex"} for B1. Thus, we select \textit{average sentence and word lengths} as a linguistic flag to capture this aspect. The full table of average values of linguistic flags can be found in Appendix~\ref{app:linguistic_flag_means}.\\

\noindent \textbf{Exemplars (\textbf{\textsc{Standardize}}-$\boldsymbol{E}$)} represent the recommended examples by experts or developers of standards for reference of users. The addition of exemplars or any artifact found in the standard that showcases gold-standard output allows the generative model to have a sense of \textit{implicit knowledge} during the content generation process. For example, in Figure~\ref{fig:exemplars}, the exemplars for a B1-level content include \textit{Frankenstein} by Mary Shelley, a well-known piece of gothic fiction. Although indirectly, any large language model trained using internet data (e.g., Wikipedia dumps) may have already formed a sense of knowledge of how this literature looks like \cite{karamolegkou-etal-2023-copyright,petroni-etal-2019-language}. We use the actual recommended exemplars from the CCS while we collected exemplars from the Penguin Readers publishing platform\footnote{\url{https://www.penguinreaders.co.uk/}} which provides expert-curated literature for CEFR. The full list of exemplars for both standards can be found in the Appendix~\ref{app:exemplars}. \\

\noindent \textbf{All (\textbf{\textsc{Standardize}}-$\star$)} represents the combination of all extract knowledge artifacts mentioned above in one prompt.

\section{Experimental Setup}
\label{sec:generation_recipe}
In this section, we detail the specifications and technical configurations for the study's main experiments. We also cover information on the datasets used, models, and generation tasks.

\subsection{Tasks and Datasets}
For this study, we specifically center our experimentation on the general task of story or narrative generation. We consider the subfield's rich literature and active research community in NLP \cite{alhussain2021automatic}, as well as being one of the most common examples demonstrated across the education community regarding the use of generative text interfaces for content generation \cite{kasneci2023chatgpt,whalen2023chatgpt}. Further, we differentiate two tasks used in our work for narrative generation as listed below. \\

\noindent \textbf{Task 1: Context Assisted Story Generation}. For this setup, we provide preliminary context in the form of $50$ to $70$ words (or approximately $3$ to $5$ sentences) in the prompt to guide the generative language model in producing the story continuation. We select the CEFR as the standard of choice to evaluate this approach and use the {European Language Grid (\textsc{ELG})} corpus\footnote{Can be accessed by filling up the form: \url{https://live.european-language-grid.eu/catalogue/corpus/9477}}\footnote{We note that the \textsc{ELG} corpus is not included in any of the pretraining data reported from the documentation of the selected generative models for experimentation, which makes it a practical option to be used in this study.} compiled by \citet{breuker2022cefr} to construct the prompts. The balanced corpus contains $300$ CEFR-aligned English texts produced by experts and distributed across five levels \textsc{A2, B1, B2, C1, C2} with $60$ instances each. A1 is omitted due to lack of resources (n < $20$). \\

\noindent \textbf{Task 2: Theme Word Story Generation}. In contrast to the previous setup, this method introduces only a single theme word for the generative language to produce a narrative from scratch, which allows for increased diversity in the content \cite{daza-etal-2016-automatic,peng-etal-2018-towards}. To compile a theme words list, we select $50$ random English noun words in plural form (e.g., \textit{dragons, mysteries, voyages}) from the {Corpus of Contemporary American English (\textsc{COCA})} \cite{davies2009385+} and prompt the generative model iteratively for each level in the standard. We investigate the application of CCS as the standard of choice in this setup.

\subsection{Models}
We select a number of generative language models $\mathcal{M_\theta}$ for content generation, each with its own advantage. For the open models, we use a number of well-known models in the 2B-7B range, including Llama2-Chat-7B \cite{touvron2023llama}, OpenChat-7B \cite{wang2023openchat}, and Longform-2.7B \cite{koksal2023longform}. For the closed model, we use GPT-4-Turbo \cite{OpenAI2023GPT4TR}. More information on the models can be found in Appendix \ref{app:models}. 

\subsection{Automatic Evaluation}
We perform a diverse set of evaluation methods $\mathcal{L}$ given examples from gold-standard datasets $\mathbf{E}$ to test the qualities of the generated content of models, as discussed further below. \\

\noindent \textbf{Model-Based Classifiers}. For the context-assisted story generation task using CEFR standards with 5 classes, we use a Random Forest classifier trained from a separate collection of Cambridge Exams dataset with CEFR labels used in the works of \citet{xia-etal-2016-text} and \citet{imperial2023flesch}. This classifier has an accuracy of $0.912$ using $79$ length-normalized\footnote{This pertains to using average-based features (e.g., the average count of sentences) in order for the classifier to avoid being confounded by total-based features (e.g., the total count of sentences).} linguistic features. For the theme word story generation using CCS standards with 2 classes, we used an XGBoost classifier from the work of \citet{imperial-2021-bert} trained from the only CCS-aligned data found online and compiled by \citet{flor-etal-2013-lexical} with an accuracy of $0.917$ using a combination of BERT embeddings and the same linguistic features stated above. Due to its limited size of $168$, we grouped the dataset into binary categories, elementary (grades $4-8$) and advanced (grades $9-12$), with $48$ and $73$ documents per class, respectively. We consider both classifiers in our work for their high accuracies ($>90\%$).\\

%For the context-assisted story generation task using CEFR standards, a Random Forest classifier is trained from a separate collection of Cambridge Exams dataset with CEFR labels following the works of \citet{xia-etal-2016-text} and \citet{imperial2023flesch}. This corpus is relatively balanced, with $331$ reading passages distributed over labels A2 to C2 (roughly $60$-$70$ each). We obtain the best accuracy with $0.912$ using $79$ length-normalized\footnote{This pertains to using average-based features (e.g., average count of sentences) in order for the classifier to avoid being confounded by total-based features (e.g., total count of sentences).} linguistic features via the LFTK tool \cite{lee-lee-2023-lftk}. For the theme word story generation using CCS standards, a binary XGBoost classifier is trained from the only CCS-aligned data found online and compiled by \citet{flor-etal-2013-lexical} with an accuracy of $0.750$ using the same feature length-normalized feature set as described above. Due to its limited size of $168$, we grouped the dataset into binary categories, easy (grades $4-8$) and intermediate (grades $9-12$), with $48$ and $73$ documents per class, respectively. We also use the Cambridge Exams \cite{xia-etal-2016-text} and the official CCS-aligned data \cite{flor-etal-2013-lexical} for computing gold-standard means mentioned in Section~\ref {sec:knowledge_artifacts}. \\
%These datasets are exclusively used for evaluation. Therefore, we do not see any data contamination risks with this setup.

\noindent \textbf{Fluency and Diversity}. We evaluate the level of fluency and content diversity of the generated content by the models as done in previous narrative generation works \cite{delucia-etal-2021-decoding,see-etal-2019-massively}. The former is measured through perplexity with an external GPT-2 model, while the latter is the density of distinct $n$-grams. \\

% Using model-based classifiers as evaluators of standard alignment may be direct but may still impose a few weaknesses, such as dependence on accuracy performance as well as quantity of training data. 
\noindent \textbf{Linguistic Similarity}. We evaluate the level of linguistic similarity of the generated content against the gold-standard datasets for CEFR (\textsc{ELG}) and CCS (\textsc{COCA}) as mentioned in Section~\ref{sec:generation_recipe}. For this method, we calculate the mean Euclidean distance of all the linguistic flags used for both standards and their levels listed in Table~\ref{tab:full_main_specs_table}. This method provides a notion of how \textit{close} the characteristics of a set of model-generated texts (e.g., GPT-4 generated B1 texts) is to its equivalent gold standard (e.g., actual B1-level texts written by experts). \\

\subsection{Expert Annotator Evaluation}
To confirm the quality of model-generated content, we also perform an evaluation using judgment from domain experts. Through our university network, we collaborated with three experts with $15-30$ years of experience in linguistic and language assessment with frameworks such as CEFR, CCS, TOEFL, and IELTS. Drawing on the methods used in previous studies \cite{delucia-etal-2021-decoding}, we asked the experts to judge the model-generated content through the following variables below. Additional information on the human evaluation can be found in Appendix~\ref{app:additional_info_evals}.\\

\noindent \textbf{Grammaticality and Coherence}. The former variable evaluates the level of \textit{naturalness} or fluency of the generated output as if it has been written by a native English speaker. The latter measures the level of cohesion between sentences where the narrative stays on-topic, and the text overall builds a \textit{consistent} story and the flow of information is smooth and easy to follow.\\

\noindent \textbf{Grade Complexity Distinction}. This variable measures the \textit{obviousness} of the complexity of a generated story on a target level (e.g., A1) with respect to another story of a different level (e.g., A2). This variable is relatively more challenging than the other metrics, as the difference between adjacent levels may not be as straightforward without referring to the quantitative characteristics of the texts. However, we included this assessment in the evaluation process to judge the quality of the model-generated texts. 

%\noindent \textbf{Alignment Metrics}. To evaluate the models' content alignment or ability to generate content based on the target reading level, we use \textbf{precise accuracy} and \textbf{adjacent accuracy}. The former is a common metric for classification tasks, while the latter preserves the ordinality by counting labels off by one class as correct. We do not apply adjacent accuracy for CCS since the labels are binary.

\begin{table*}[!t]
\scriptsize
\centering

\begin{minipage}[h]{.5\textwidth}
\centering
\setlength\tabcolsep{3pt}
\begin{tabular}{@{}lrrrr@{}}
\toprule
\textbf{Model} &
  \begin{tabular}[c]{@{}r@{}}\bf Precise\\ \bf Accuracy\end{tabular} &
  \begin{tabular}[c]{@{}r@{}}\bf Adjacent\\ \bf Accuracy\end{tabular} &
  \begin{tabular}[c]{@{}r@{}}\bf Fluency\\ \emph{(perplexity)}\end{tabular} &
  \begin{tabular}[c]{@{}r@{}}\bf Diversity\\ \emph{(distinct-n)}\end{tabular} \\
\midrule
\rowcolor{gray!15}
\textbf{Llama2 7B} &
  \multicolumn{1}{l}{} &
  \multicolumn{1}{l}{} &
  \multicolumn{1}{l}{} &
  \multicolumn{1}{l}{} \\
- Teacher Style              & 0.203 & 0.636 & \bf {13.189 $\pm$4.88}     & 0.156 $\pm$0.03        \\
- \textsc{Standardize-$A$}           & 0.270 & 0.626 & 13.694 $\pm$7.74        & 0.155 $\pm$0.02        \\
- \textsc{Standardize-$E$}        & 0.320 & \bf 0.683 & 15.576 $\pm$3.31      & 0.188 $\pm$0.01        \\
- \textsc{Standardize-$L$} & 0.273 & 0.606 & 20.175 $\pm$4.47        & 0.186 $\pm$0.01        \\ 
- \textsc{Standardize-$\star$} & \bf 0.354 & 0.670 & 17.892 $\pm$3.94        & \bf 0.193 $\pm$0.01        \\ \midrule

\rowcolor{gray!15}\textbf{OpenChat 7B}         &       &     & \multicolumn{1}{l}{} & \multicolumn{1}{l}{} \\
- Teacher Style              & 0.237 & 0.626 & 22.039 $\pm$7.70       & 0.170 $\pm$0.02        \\
- \textsc{Standardize-$A$}           & 0.243 & \bf 0.630 & 21.195 $\pm$7.66        & 0.171 $\pm$0.02        \\
- \textsc{Standardize-$E$}        & 0.253 & 0.600 & \bf 13.931 $\pm$2.97     & 0.178 $\pm$0.01        \\
- \textsc{Standardize-$L$} & \bf 0.270 & 0.546 & 18.182 $\pm$8.52  & \bf 0.179 $\pm$0.02        \\
- \textsc{Standardize-$\star$} & 0.253 & 0.596 & 12.806 $\pm$2.70        & 0.171 $\pm$0.03        \\\midrule

\rowcolor{gray!15}\textbf{Longform 3B}         &       &    & \multicolumn{1}{l}{} & \multicolumn{1}{l}{} \\
- Teacher Style              & 0.230 & 0.606 & 18.209 $\pm$6.01       & 0.159 $\pm$0.02        \\
- \textsc{Standardize-$A$}           & 0.223 & \bf 0.610 & 17.982 $\pm$9.21        & 0.157 $\pm$0.02        \\
- \textsc{Standardize-$E$}        & 0.257 & 0.496 & 25.075 $\pm$8.80       & \bf 0.192 $\pm$0.11        \\
- \textsc{Standardize-$L$} & \bf 0.283 & 0.586 & \bf 16.926 $\pm$6.91     & 0.161 $\pm$0.03        \\
- \textsc{Standardize-$\star$} & 0.277 & 0.543 & 16.806 $\pm$7.40        & 0.170 $\pm$0.04 \\ \midrule

\rowcolor{gray!15}\textbf{GPT-4}               &       &    & \multicolumn{1}{l}{} & \multicolumn{1}{l}{} \\
- Teacher Style              & 0.227 & 0.630 & 27.357 $\pm$6.30      & 0.187 $\pm$0.08        \\
- \textsc{Standardize-$A$}           & 0.397 & 0.846 & 29.729 $\pm$9.58     & 0.174 $\pm$0.01        \\
- \textsc{Standardize-$E$}        & 0.307 & 0.703 & 30.357 $\pm$9.79        & 0.182 $\pm$0.01        \\
- \textsc{Standardize-$L$} & {0.480} &  \bf {0.906} &  24.115 $\pm$7.04       & {0.194 $\pm$0.03}\\ 
- \textsc{Standardize-$\star$} & \bf 0.540 & 0.803 & \bf 22.591 $\pm$1.61        & \bf 0.218 $\pm$0.05 \\
\bottomrule    
\end{tabular}
\caption{Experiment results comparing the conventional teacher style prompting with the \textsc{Standardize} framework for the Common European Framework of Reference for Languages (CEFR) standards. }
\label{tab:cefr_main_table}
\end{minipage}
\hfill
\begin{minipage}[h]{.45\textwidth}
\centering
\setlength\tabcolsep{3pt}
\begin{tabular}{@{}lrrr@{}}
\toprule
\textbf{Model} &
  \begin{tabular}[c]{@{}r@{}}\bf Precise\\ \bf Accuracy\end{tabular} &
  \begin{tabular}[c]{@{}r@{}}\bf Fluency\\ (\emph{perplexity})\end{tabular} &
  \begin{tabular}[c]{@{}r@{}}\bf Diversity\\ (\emph{distinct-n})\end{tabular} \\ \midrule
\rowcolor{gray!15}\textbf{Llama2 7B} &
  \multicolumn{1}{l}{} &
  \multicolumn{1}{l}{} &
  \multicolumn{1}{l}{} \\
- Teacher Style              & 0.470 & 17.936 $\pm$4.32       & 0.184 $\pm$0.01        \\
- \textsc{Standardize-$A$}           & 0.580 & 22.070 $\pm$1.75       & 0.171 $\pm$0.01        \\
- \textsc{Standardize-$E$}        & 0.570 & \bf 13.484 $\pm$2.50       & \bf 0.193 $\pm$0.01        \\
- \textsc{Standardize-$L$} & \bf {0.720} & 15.066 $\pm$2.47       & 0.191 $\pm$0.01        \\
- \textsc{Standardize-$\star$} & 0.623 & 14.707 $\pm$2.40       & \bf 0.193 $\pm$0.01        \\ \midrule

\rowcolor{gray!15}\textbf{OpenChat 7B}         &       & \multicolumn{1}{l}{} & \multicolumn{1}{l}{} \\
- Teacher Style              & 0.470 & 16.116 $\pm$12.39      & 0.166 $\pm$0.05        \\
- \textsc{Standardize-$A$}           & 0.550 & 19.444 $\pm$2.57       & 0.172 $\pm$0.01        \\
- \textsc{Standardize-$E$}        & 0.490 & {12.438 $\pm$1.85}       & 0.178 $\pm$0.01        \\
- \textsc{Standardize-$L$} & \bf 0.580 & 13.734 $\pm$2.53       & \bf 0.180 $\pm$0.01        \\ 
- \textsc{Standardize-$\star$} & 0.560 & \bf 10.717 $\pm$1.53       &  0.169 $\pm$0.01 \\ \midrule

\rowcolor{gray!15}\textbf{Longform 3B}         &       & \multicolumn{1}{l}{} & \multicolumn{1}{l}{} \\
- Teacher Style              & 0.500 & 13.657 $\pm$5.39       &  0.154 $\pm$0.04        \\
- \textsc{Standardize-$A$}           & 0.450 & 17.918 $\pm$4.74       & 0.148 $\pm$0.01        \\
- \textsc{Standardize-$E$}        & 0.510 & 14.277 $\pm$2.79       & 0.151 $\pm$0.02        \\
- \textsc{Standardize-$L$} & 0.610 & 13.398 $\pm$3.93       & 0.148 $\pm$0.04        \\ 
- \textsc{Standardize-$\star$} & \bf 0.620 &  \bf 10.400 $\pm$1.53       &  \bf 0.169 $\pm$0.01 \\ \midrule

\rowcolor{gray!15}\textbf{GPT-4}               &       & \multicolumn{1}{l}{} & \multicolumn{1}{l}{} \\
- Teacher Style              & 0.590 & 32.447 $\pm$7.46       &  {0.195 $\pm$0.01}        \\
- \textsc{Standardize-$A$}           & 0.550 & 31.765 $\pm$11.30      & 0.169 $\pm$0.01        \\
- \textsc{Standardize-$E$}        & 0.520 & 29.912 $\pm$6.81      & 0.184 $\pm$0.01        \\
- \textsc{Standardize-$L$} &  0.610 & 26.912 $\pm$6.11       & 0.155 $\pm$0.01       \\
- \textsc{Standardize-$\star$} & \bf 0.790 & \bf 21.277 $\pm$4.50       &  \bf 0.198 $\pm$0.01 \\
\bottomrule
\end{tabular}
\caption{Experiment results comparing the conventional teacher style prompting with the \textsc{Standardize} framework for the Common Core Standards (CCS). }
\label{tab:ccs_main_table}
\end{minipage}
\end{table*}

\section{Results and Discussion}
\label{sec:results}
We discuss the results of our experiments procedures with the methods from the \textsc{Standardize} framework.

\subsection{Standard Alignment via Classification Performance}
The overall performance of models for CEFR and CCS are reported in Tables~\ref{tab:cefr_main_table} and~\ref{tab:ccs_main_table}. For CEFR, the top-performing setup across the four models all belong to the \textsc{Standardize} framework. We report over a $100$\% increase in performance using the best setup with GPT-4 with \textsc{Standardize}-$\star$ in precise accuracy from $0.227$ to $0.540$ and a $43$\% increase for adjacent accuracy from $0.630$ to $0.906$ compared to the teacher style method. Through \textsc{Standardize}, open models also gained substantial boosts in performance, such as Longform up by $23$\%, OpenChat up by $14$\%, and Llama2 by $74$\%. In terms of adjacent accuracies, GPT-4 remained the best model for preserving the ordinality of the labels with $0.906$, up by $44\%$. With CCS, the general scores obtained in this setup are higher compared to CEFR with five classes due to binary labeling. We see a similar pattern where all open and closed models obtained the best performance, with boosts ranging from $3$\% to $45$\% using linguistic flags \textsc{Standardize}-$L$ and a combination of all knowledge artifacts \textsc{Standardize}-$\star$ to refine the generated content toward the target level. From these findings, \textbf{we provide concrete evidence that using the actual content of the standards through knowledge artifact representations from \textsc{Standardize} may be crucial} when prompting LLMs via in-context learning to produce standard-aligned content for classroom use. 

%We note that the Llama2 model ($0.637$) was able to surpass the performance with GPT-4 ($0.610$) by a margin in this setup. We see it as a good opportunity for furthering the research efforts for open models in tasks like these, as it shows that open models do have the potential to perform better in certain tasks compared to closed models.

\subsection{Standard Alignment via Linguistic Similarity} 
We visualize the distributions of the best performing \textsc{Standardize} methods in Figures~\ref{fig:ave_sen_len} to~\ref{fig:corr_ttr} with comparison to the teacher style method. From the results, we observe that the general trend of using \textsc{Standardize} produces a more \textit{stable} distribution across the variables it is explicitly controlling for (e.g., \textit{average sentence length} or \textit{type token diversity} as listed in Table~\ref{tab:full_main_specs_table}), particularly with the CCS standards. We also notice that the distributions using \textsc{Standardize}-$L$ also produce distributions closer to the mean (represented as a yellow star) from their corresponding gold-standard data. Moreover, in terms of linguistic similarity, as reported in Table~\ref{tab:euclidean_dist}, \textsc{Standardize} makes the quality of model generations \textit{more similar} to the linguistic characteristics of the gold standard datasets in CEFR and CCS. Overall, \textbf{these findings further strengthen the evidence of using \textsc{Standardize} in producing linguistically similar content with gold-standard data} compared to the conventional teacher style method.

\begin{table}[!htbp]
\footnotesize
\centering
\begin{subtable}{\linewidth}
\setlength\tabcolsep{4pt}
\begin{tabular}{@{}l|rrrrr@{}}

{Setup} & \multicolumn{1}{c}{A2} & \multicolumn{1}{c}{B1} & \multicolumn{1}{c}{B2} & \multicolumn{1}{c}{C1} & \multicolumn{1}{c}{C2} \\ \midrule
Teacher Style  & 136.7                  & \bf 96.7                   & 169.9                  & 307.3                  & 291.6                  \\
\bf \textsc{Standardize}-$\star$  & \bf 61.4                   & 106.2                  & \bf 97.64                  & \bf 219.6                  & \bf 234.7                  %\bottomrule
\end{tabular}
\end{subtable}

\vspace{10pt}

\begin{subtable}{\linewidth}
\setlength\tabcolsep{4pt}
\begin{tabular}{@{}l|rr@{}} 

{Setup} & \multicolumn{1}{c}{Elementary} & \multicolumn{1}{c}{Advanced}  \\\midrule
Teacher Style  & 76.1                    & 157.9                    \\
\bf \textsc{Standardize}-$\star$  & \bf 63.8                    & \bf 125.7          
\end{tabular}
\end{subtable}

\caption{Mean Euclidean distances of generated content using simple teacher style prompting vs. \textsc{Standardize}-$\star$ for CEFR (top) and CCS (bottom).}
\label{tab:euclidean_dist}
\end{table}

%%
%% EXPERT EVALS
%%
\begin{figure*}[!t]
    \centering
    \begin{subfigure}[t]{0.34\linewidth}
        \centering
        \includegraphics[width=\linewidth]{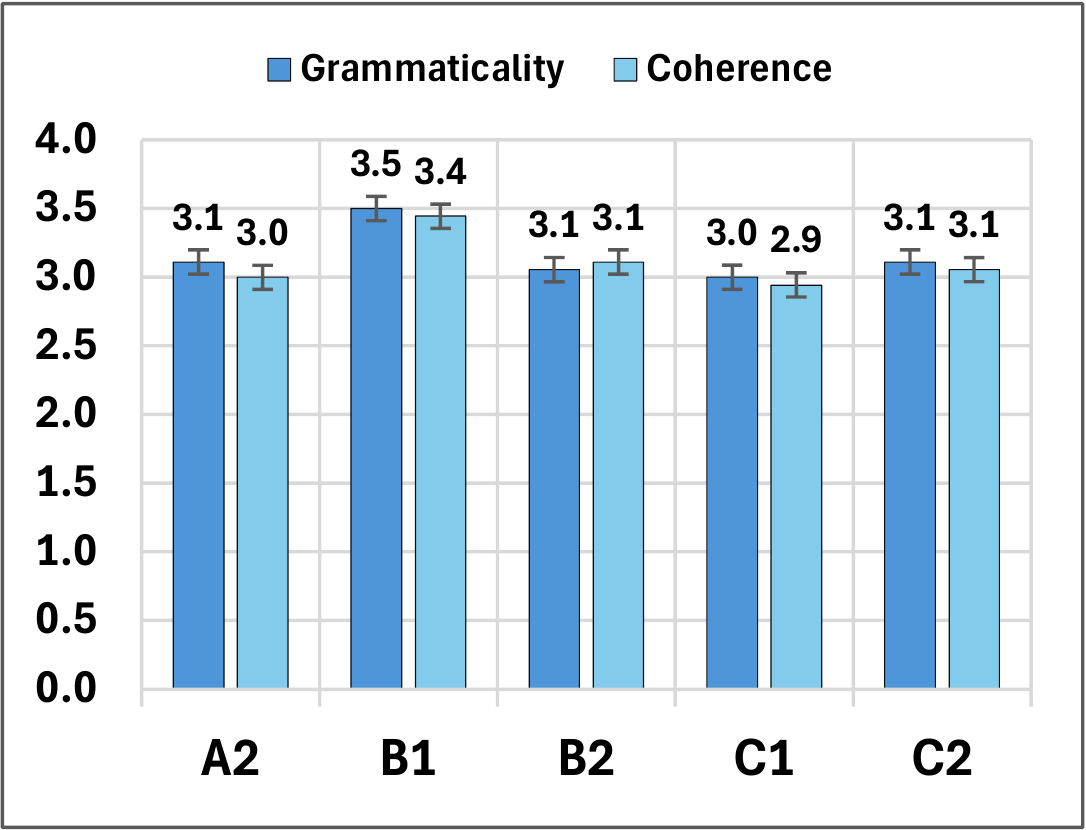}
        \caption{Expert evaluation on the generation quality of the GPT-4 model with \textsc{Standardize}-$\star$ for CEFR. Inter-rater reliability using Kendall's $W$ is $0.34$ which denotes moderate agreement.}
        \label{fig:cefr_expert_evaluation}
    \end{subfigure}
    \hfill
    \begin{subfigure}[t]{0.275\linewidth}
        \centering
        \includegraphics[width=\linewidth]{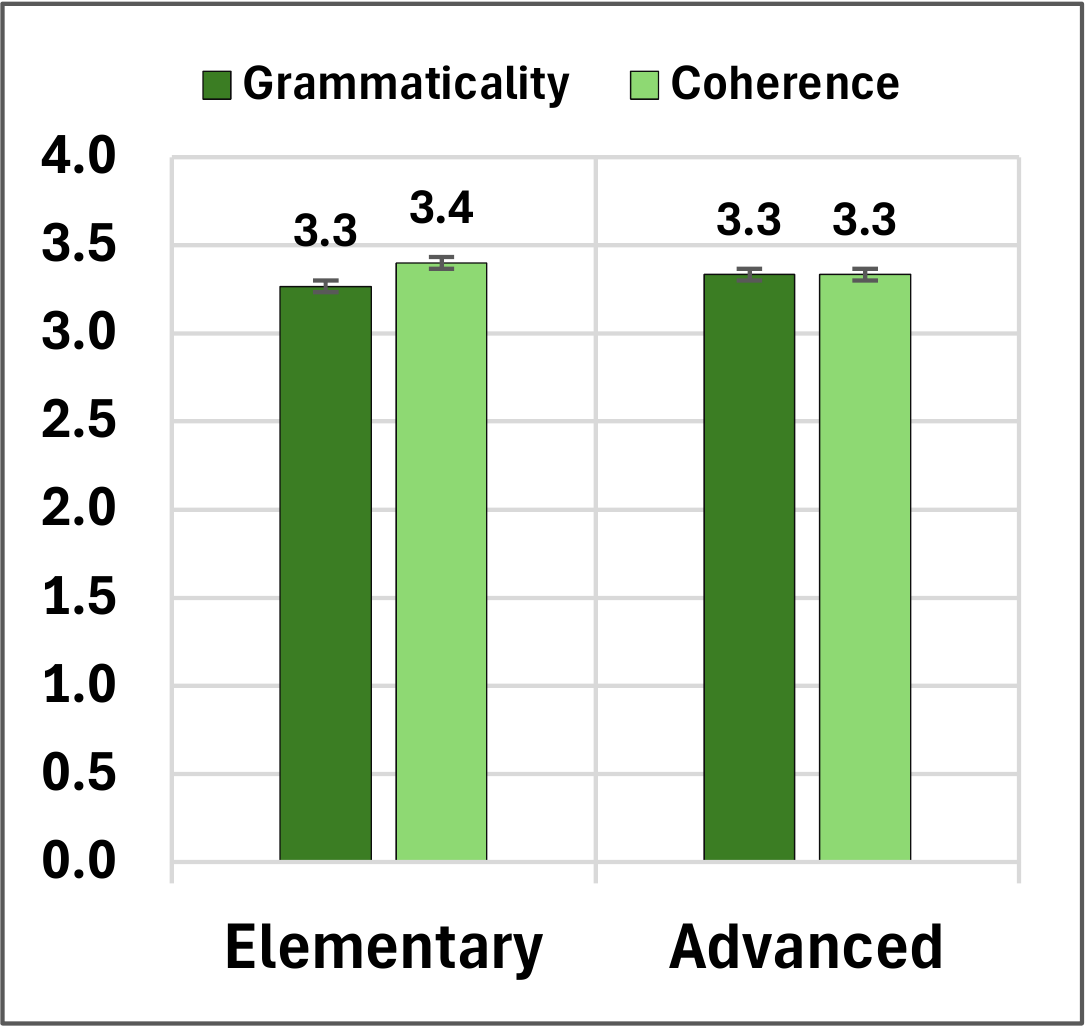}
        \caption{Expert evaluation on the generation quality of the GPT-4 model with \textsc{Standardize}-$\star$ for CCS. Inter-rater reliability using Kendall's $W$ is $0.40$ which denotes strong agreement}
        \label{fig:ccs_expert_evaluation}
    \end{subfigure}
    \hfill
    \begin{subfigure}[t]{0.32\linewidth}
        \centering
        \includegraphics[width=\linewidth]{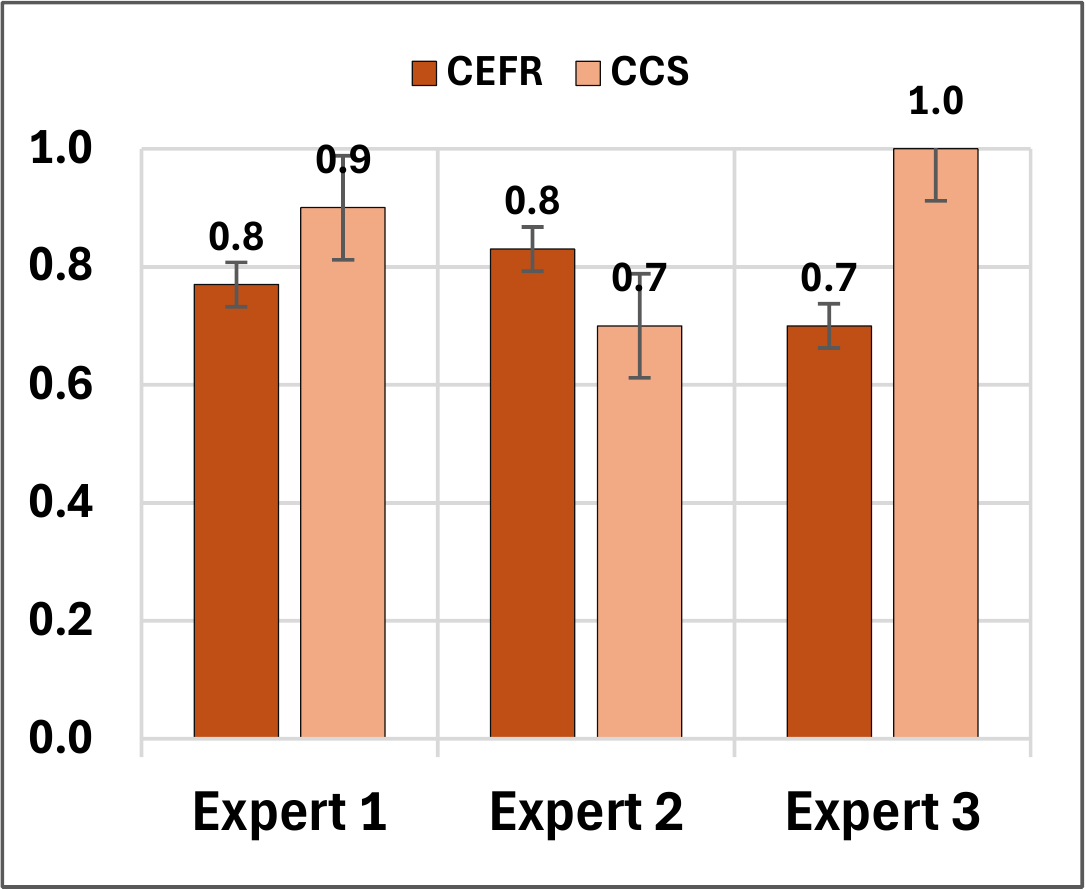}
        \caption{Performance of expert evaluators on estimating the complexity of generated content for CEFR and CCS. Inter-rater reliability using Kendall's $W$ is $0.45$ which denotes strong agreement.}
        \label{fig:grade_distinction}
    \end{subfigure}
    \hfill
    \caption{Overview of mean ratings of grammaticality or fluency, coherence, and grade complexity distinction from the human expert evaluations using the top-performing models for CEFR and CCS. All evaluation procedures obtain generally favorable results as well as acceptable inter-rater reliability scores (equal and above the threshold of $0.30$)}.
    \label{fig:expert_evaluation_results}
\end{figure*}

\subsection{Assessment of Generation Qualities via Expert Judgment and Automatic Metrics} 
For both computed fluency and content diversity, we see similar results from the previous evaluation techniques where the best performing models are all models improved through the \textsc{Standardize} framework particularly OpenChat, Longform, and GPT-4. Looking at expert evaluations as reported in Figure~\ref{fig:expert_evaluation_results}, we observe consistent high ratings on grammaticality and coherence of the topi performing model, GPT-4 with \textsc{Standardize}-$\star$, for both CEFR and CCS with an average of $3.13$ and $3.35$, respectively. On the grade complexity distinction, all three expert evaluators were able to achieve high accuracies ($>0.70$) in selecting correct simple and complex texts from the model-generated data, denoting the obviousness of complexity. Likewise, all expert evaluation tests achieved strong inter-rater reliability scores ($>0.30$) through Kendall's $W$ \cite{kendall1948rank}. With these findings, \textbf{we affirm the effectivity of the \textsc{Standardize} framework through expert judgment on generating more fluent, grammatical, grade-distinct, and diverse content compared to the teacher-style approach}.

%Our experiment procedure is focused on generating text content that aligns with the specified target level with respect to a standard. The standards that we applied in this study, CEFR and CCS, did not explicitly provide information on content creativity and how to measure this. Thus, we posit that improving creativity may be an interesting angle to explore in future works. 

%But this may be on a case-to-case basis and task-dependent since we do not see the same tradeoff in performance with the CEFR standards in context-assisted story generation. 
%While it may not be a surprise that these LLMs are capable of generating fluent and human-readable texts due to their high parameter count in billions (2.7B - 7B), we still see an advantage of the performance boost provided by incorporating the knowledge artifacts from the standards for this task, particularly with the linguistic flags.

\section{Implications to Generative Models for Education}
\label{sec:discussion}

We discuss important points highlighting the real-world implications of our study within and beyond language model experimentations.\\

\noindent \textbf{Validity on Global Education Context}. Our main contribution, the \textsc{Standardize} framework, leverages the idea of a more holistic method for capturing the intricacies and complexities of educational standards for content generation. Our experiments with the CEFR and CCS standards showcase an opportunity for the generated texts of language model interfaces such as GPT-4, which are commonly used by educators and teachers, to be aligned with international language proficiency levels. Moreover, showing the effectiveness of \textsc{Standardize} on the aforementioned internationally recognized academic standards used in European and Northern American schools signifies the framework's strong potential for cross-curricula application. Thus, we invite future researchers to explore, validate, and propose derivations of our base framework for their own languages and language-specific standards for content generation.\\

\noindent \textbf{Towards More Personalized Content Generation}. Investigating the potential of generative models for personalized learning, such as providing adaptive feedback aligned with students' needs, is an active area in AI for education \cite{kasneci2023chatgpt,meyer2023chatgpt,sailer2023adaptive,tack2022ai}. This work contributes toward the goal of helping educators craft more personalized content for learners using the capabilities of large language models based on an assigned language proficiency level described by a standard. While we present a novel task specifically targeted for the NLP community to encourage research in this direction (\textsc{Standard-CTG} as covered in Section~\ref{sec:standard-ctg}), our results may be useful for educators by providing context on better methods for generating level or target audience-specific texts by prompting language models using information found in educational standards.

\section{Related Work}
Research in complexity-controlled generation has explored diverse variables in terms of text format, granularity, and task variation. The work of \citet{agrawal-carpuat-2019-controlling} introduced controlling for specific complexity in the machine translation task. The following works of \citet{agrawal-carpuat-2023-controlling} and \citet{ribeiro-etal-2023-generating} explored grade-specific text simplification and summarization using control tokens and reinforcement learning, respectively. Currently, only two works have investigated incorporating CEFR for language learning content generation. \citet{stowe-etal-2022-controlled} and \citet{imperial2023flesch} both made use of CEFR-aligned text for NLG. However, none of them made use of the actual guideline information found in CEFR during the generation process. 

Our study's main novelty is the holistic capture of expert-defined standards by exploring possible representations we call artifacts that can improve how a language model refines its content generation process with respect to a target language proficiency level. We emphasize the importance of the use of in-context learning without additional finetuning in this work to preserve the capabilities of models across other language-related tasks. Our \textsc{Standardize} framework derives motivation from \citet{zhou2023controlled} and \citet{ram-etal-2023-context}, where a verbalizer is used to transform quantitative constraints into natural language for prompting, as well as the use of a lookup and retrieval phase where aspect information is added in the prompt to influence model controllability. 

%In comparison to all the works mentioned, our study's main novelty is the holistic capture of expert-defined standards by exploring possible representations we call artifacts that can improve how a language model refines its content generation process with respect to a target language proficiency level.

\section{Conclusion}
In this work, we proposed the \textsc{Standardize} framework using knowledge artifacts that allowed large language models such as Llama2 and GPT-4 to gain significant performance boosts ($45$\% - $100$\%) on generating content aligned with educational standards as well as preserving important narrative qualities such as fluency, grammaticality, coherence, and grade distinctness. From this, we see a very promising potential for cross-domain and cross-standard generalization of our proposed method with the range of educational contexts around the world and invite future work to build on our baseline models.

%Our study has two contributions: (a) \textsc{Standard-CTG}, a new task formalizing the use of complex constraints of a standard as a reference for control, and (b) \textsc{Standardize}, which is a novel framework for solving the task by using extract knowledge artifacts. Our findings provide strong evidences that incorporating \textsc{Standardize} to generative models provides significant boosts in the ability to generate standard-aligned content, with the best model GPT-4 obtaining double the performance of baseline models.

\section*{Ethical Considerations}
All datasets and corpora used in this study, such as the \textsc{ELG} \cite{breuker2022cefr}, Cambridge Exams \cite{xia-etal-2016-text}, and CCS \cite{flor-etal-2013-lexical}, are already established and accessible for research purposes. We observe a specific tone in the discussion of our experiments, emphasizing that the main motivation of the work is that language models such as GPT-4 can provide assistance in producing content that is more aligned or faithful with the constraints of standards such as CEFR or CCS {without} implying that they can replace experts in the field or produce better quality than the gold-standard data. Further, we also {do not} imply that any model enriched by any computational method to produce more standard-aligned content can replace the standard itself. Overall, we do not foresee any serious ethical issues in this study. 

\section*{Limitations}

\noindent\textbf{Language Coverage of Standards}. This work is mainly centered on the use of datasets and standards for the English language. While standards for language assessment, such as CEFR, have expanded through the years with versions to cover other languages, such as German, Czech, and Italian \cite{vajjala-rama-2018-experiments}, we do not claim that our results will be able to generalize and have the same advantages with these languages. However, investigating this direction may be a good research opportunity for future work.\\

\noindent\textbf{Dependence on Evaluation Methods}. As observed in Section~\ref{sec:results}, we made sure to cover a variety of evaluation procedures for testing standard alignment instead of only using model-based methods such as a classifier. The limitation here is that trained classifiers are dependent on factors such as their accuracy, the quantity of data, the complexity of the training algorithm, and the quality of features. Thus, other means of evaluating alignment that is more direct, such as computed feature distances against a gold-standard dataset, is always recommended. Moreover, our model-based CEFR and CCS evaluators make use of artifacts such as datasets and tools for feature extraction from peer-reviewed papers \cite{xia-etal-2016-text,flor-etal-2013-lexical}. We are aware of paid third-party services online that promise more accurate classification of labels in CEFR, but they generally do not provide details on linguistic predictors used for prediction. Thus, this may not be a practical option for research. \\

\noindent\textbf{Attribute-Based Standards}. The standards used in this study, CEFR and CCS, are attribute-based standards that specify recommended characteristics of texts that are countable (e.g., sentence length or average number of words). These specifications contribute towards the overall complexity of texts which are within the scope of CEFR and CCS. Standards in other domains may come in different forms of constraints, such as dependence on an external specialized vocabulary or following specific sequential processes to arrive at a result. Moreover, our exploration of CEFR and CCS standards is centered on the downstream task of narrative generation, as this fits the most generic form of reading material in classrooms. We leave the exploration of extending the \textsc{Standardize} framework to other domains that also observe attribute-based specifications as well as other adjacent text generation tasks (e.g., summary generation) in future work.

\section*{Acknowledgements}
We are grateful to the anonymous reviewers and Action Editors in ARR for their feedback on the improvement of this paper and to Dr. Brian North for the insightful discussions on capturing language standards, including CEFR, as part of the theoretical component of this work. We also thank Dr. Samantha Curle and Dr. Reka Jablonkai from the Department of Education at the University of Bath for helping with the evaluation of model-generated texts. This work made use of the Hex GPU cloud of the Department of Computer Science at the University of Bath. JMI is supported by the National University Philippines and the UKRI Centre for Doctoral Training in Accountable, Responsible, and Transparent AI [EP/S023437/1] of the University of Bath. We attribute the black icons used in Figure~\ref{fig:fig1} to the collections of Design Circle and Victor Zukeran from the Noun Project and the colored teacher icon from Flaticon.

% Noun ProjectProject(CC BY 3.0)
% "Download" by Victor Akio Zukeran from the Noun Project (CC BY 3.0)
% Content Search by Design Circle from the Noun Project (CC BY 3.0)

\bibliography{references,acl_references}
\bibliographystyle{acl_natbib}

\clearpage
\appendix

\section{Appendix}
\label{sec:appendix}

\subsection{Libraries and Dependencies}
We have used the following dependencies and Python libraries for the study: Linguistic Feature Tool Kit (LFTK) \cite{lee-lee-2023-lftk}, Spacy (\url{https://spacy.io/}), Scikit-Learn (\url{https://scikit-learn.org/stable/}), OpenAI API (\url{https://openai.com/blog/openai-api}).

\subsection{Corpus Statistics}
\label{app:corpus_statistics}
We provide basic statistical information about the various corpora used in the study.

\begin{table}[!htbp]
\centering
\small
\begin{tabular}{@{}llrr}
\toprule
\bf Level &
  \bf Size &
  \multicolumn{1}{r}{\begin{tabular}[c]{@{}r@{}} \bf Average  \\ \bf Word Count\end{tabular}} &
  \multicolumn{1}{r}{\begin{tabular}[c]{@{}r@{}}\bf Average\\ \bf Sentence Count\end{tabular}} \\ \midrule
A2              & 60            & 186.55          & 18.91                    \\
B1              & 60            & 264.25          & 15.90                     \\
B2              & 60            & 517.71          & 31.71                  \\
C1              & 60            & 728.93          & 40.70                   \\
C2              & 60            & 749.73          & 37.55            
\\
\bottomrule
\end{tabular}
\caption{Statistics of the \textsc{ELG} corpus \cite{breuker2022cefr} used for the CEFR context assisted story generation task.}
\label{tab:elg_stats}
\end{table}

\begin{table}[!htbp]
\centering
\small
\begin{tabular}{@{}llrr}
\toprule
\bf Grade &
  \bf Size &
  \multicolumn{1}{r}{\begin{tabular}[c]{@{}r@{}} \bf Average  \\ \bf Word Count\end{tabular}} &
  \multicolumn{1}{r}{\begin{tabular}[c]{@{}r@{}}\bf Average\\ \bf Sentence Count\end{tabular}} \\ \midrule
Elementary              & 48            & 204.91          & 28.55                    \\
Advanced              & 73            & 255.17          & 31.08                             
\\
\bottomrule
\end{tabular}
\caption{Statistics of the official \textsc{CCS}-aligned corpus \cite{flor-etal-2013-lexical} used as gold-standard dataset for the \textsc{Standardize}-$L$ artifact and for training the CCS classifier used in Section~\ref{sec:results}.}
\label{tab:ccs_stats}
\end{table}

\begin{table}[!htbp]
\centering
\small
\begin{tabular}{@{}llrr}
\toprule
\bf Level &
  \bf Size &
  \multicolumn{1}{r}{\begin{tabular}[c]{@{}r@{}} \bf Average  \\ \bf Word Count\end{tabular}} &
  \multicolumn{1}{r}{\begin{tabular}[c]{@{}r@{}}\bf Average\\ \bf Sentence Count\end{tabular}} \\ \midrule
A2              & 64            & 60.87          & 11.53                    \\
B1              & 60            & 122.38          & 16.25                     \\
B2              & 71            & 265.35         & 37.03                  \\
C1              & 67            & 355.71          & 43.37                   \\
C2              & 69            & 333.86         & 38.41            
\\
\bottomrule
\end{tabular}
\caption{Statistics of the Cambridge Exams corpus \cite{xia-etal-2016-text} used as gold-standard dataset for the \textsc{Standardize}-$L$ artifact and for training the CEFR classifier used in Section~\ref{sec:results}.}
\label{tab:cambridge_exams_stats}
\end{table}

\subsection{Additional Information on Models and Inference}
\label{app:models}
We set the minimum generated new tokens to $30$ and the maximum to $300$, as well as set the nucleus sampling decoding (top-p) to $0.95$ as done with previous works on story generation \cite{imperial-madabushi-2023-uniform,delucia-etal-2021-decoding,see-etal-2019-massively}. The actual sizes of the open models range from 5GB to 15 GB max. We used a hosted GPU cloud with 4 NVIDIA Ti 3090 with 24GB memory size for model inference.\\

\noindent \textbf{Llama2-Chat} \cite{touvron2023llama2} is one of the community-recognized open instruction-tuned models released by Meta and an improved version of Llama 1 \cite{touvron2023llama}. For this task, we use the 7B version\footnote{\url{https://huggingface.co/meta-llama/Llama-2-7b-chat-hf}} finetuned from over a million human preference data and optimized for chat and dialogue use cases. We prioritized the addition of this model in our study for its accessibility to the general NLP community.\\

\noindent \textbf{Longform-OPT} \cite{koksal2023longform} is a recent instruction-tuned model optimized for long text generation using the LongForm dataset. For this study, we use the OPT model variant\footnote{\url{https://huggingface.co/akoksal/LongForm-OPT-2.7B}} \cite{zhang2022opt} with 2.7B parameters as this version obtained the best performance for the short story generation task using the \textsc{WritingPrompts} dataset \cite{fan-etal-2018-hierarchical} against other instruction-tuned models such as Alpaca-LLaMA \cite{taori2023alpaca}, FlanT5 \cite{chung2022scaling}, Tk-Instruct \cite{wang2022benchmarking}, and T0++ \cite{sanh2021multitask}. \\

\noindent \textbf{OpenChat} \cite{wang2023openchat} is the most recent open model in our experiment setup, which currently is reported to be the best 7B model as of this writing and outperforms closed models such as ChatGPT (March) across a number of benchmark tasks such as GSM8K and TruthfulQA. In contrast to Llama and GPT models, which used RLHF \cite{ouyang2022training}, OpenChat is trained with mixed-quality data which is composed of high-quality expert data and sub-optimal web data with no preference labels. We use the 7B version\footnote{\url{https://huggingface.co/openchat/openchat-3.5-0106}} of this model variant released in January 2024.\\

\noindent \textbf{GPT-4} \cite{OpenAI2023GPT4TR} is the only closed model included in this study. We decide to add this model to our experiment for its global recognition through its easy-to-use interface among interdisciplinary fields, particularly in education \cite{kasneci2023chatgpt}. We use the version\footnote{\url{https://platform.openai.com/docs/models/gpt-4-and-gpt-4-turbo}} finetuned with proprietary training data up to April 2023 with a $128$K context window.

\subsection{Exemplars List}
\label{app:exemplars}
We list the actual list of literary exemplars used for the \textsc{Standardize} framework. We manually selected at most three classical exemplars as reference for the language models.

\begin{table}[!htbp]
\centering
\footnotesize
\begin{tabular}{@{}ll@{}}
\toprule
\textbf{Level} & \multicolumn{1}{c}{\textbf{Exemplars}}            \\ \midrule
A2             & \textit{A Christmas Carol} by Charles Dickens             \\
               & \textit{The Adventures Of Huckleberry Finn} by Mark Twain \\
               & \textit{The Little Prince} by Antoine de Saint-Exupery    \\ \midrule
B1             & \textit{Frankenstein} by Mary Shelley                     \\
               & \textit{Wuthering Heights} by Emily Bronte                \\
               & \textit{Midsummer Night's Dream} by Shakespeare           \\ \midrule
B2             & \textit{Moby Dick} by Herman Melville                     \\
               & \textit{Jane Eyre} by Charlotte Bronte                    \\
               & \textit{Sense and Sensibility} by Jane Austen             \\ \midrule
C1             & \textit{Animal Farm} by George Orwell                     \\
               & \textit{Anna Karenina} by Leo Tolstoy                     \\
               & \textit{Great Expectations} by Charles Dickens            \\ \midrule
C2             & \textit{Oliver Twist} by Charles Dickens                  \\
               & \textit{Crime and Punishment} by Fyodor Dostoevsky        \\
               & \textit{Les Miserables} by Victor Hugo                    \\ \bottomrule
\end{tabular}
\caption{The full exemplar list used for CEFR standards obtained from the Penguin Reader website (\url{https://www.penguinreaders.co.uk/}).}
\end{table}

\begin{table}[!htbp]
\centering
\footnotesize
\begin{tabular}{@{}ll@{}}
\toprule
\textbf{Grade} & \multicolumn{1}{c}{\textbf{Exemplars}}            \\ \midrule
Elementary             & \textit{Little Women} by Louisa May Alcott             \\
               & \textit{The Adventures of Tom Sawyer} by Mark Twain \\
               & \textit{The Road Not Taken} by Robert Frost    \\ \midrule
Advanced             & \textit{Jane Eyre} by Charlotte Brontë                    \\
               & \textit{The Great Gatsby} by F. Scott Fitzgerald             \\
               & \textit{Fahrenheit 451} by Ray Bradbury           \\ \bottomrule
\end{tabular}
\caption{The full exemplar list used for CCS standards obtained from the official website (\url{https://www.thecorestandards.org/ELA-Literacy/}).}
\end{table}

\subsection{Mean Values of Linguistic Flags}
\label{app:linguistic_flag_means}
We provide the computed averages of the linguistic flags from the aspects of the two standards, CEFR and CCS, used in this work reported in Tables~\ref{tab:cefr_average_ling_flags} and ~\ref{tab:ccs_average_ling_flags}.

\subsection{Additional Information on Human Expert Evaluation}
\label{app:additional_info_evals}
We created and distributed the evaluation instrument through QuestionPro (\url{https://www.questionpro.com/}). In contrast to non-expert validation techniques where all instances are distributed automatically to available annotator platforms such as Amazon Turk, we use a representative random sample of our data for evaluation in consideration with the experts' time constraints. For all tests, we randomly sampled $10\%$ of the total generated narrative content using the best-performing model, which is both the GPT-4 model with \textsc{Standardize}-$\star$, for each corresponding task associated with CEFR and CCS as described in Section~\ref{sec:generation_recipe}. 

For grammaticality and coherence evaluation, we adapted the same four-point Likert scale from the work of \citet{delucia-etal-2021-decoding} for evaluating select model-generated content found through this link: \url{https://github.com/JHU-CLSP/gpt2-narrative-decoding/}. Snapshots of the instruction and test instances presented to experts for evaluation can be viewed in Figures~\ref{fig:part_A_instruction} and ~\ref{fig:part_A_example}.

For the grade complexity distinction, we adapted a simpler select-one response type where for each test instance being evaluated, we select a random test instance from the adjacent next level of the target test instance and ask the experts to select which two examples of model-generated content are more \textit{simpler} or \textit{complex}. The idea here is that the expert should be able to tell the \textit{obviousness} of the complexity of the test instance by indicating which is simpler or more complex. Snapshots of the instruction and test instances presented to experts for evaluation can be viewed in Figures~\ref{fig:part_B_instruction} and ~\ref{fig:part_B_example}.

Overall, our human evaluation design has been validated by the experts in language assessment we collaborated with through preliminary discussions on the scope, instrument, target outcomes, and presentation of the results from the task. As a form of compensation, we offered £$30$ upon completion of the entire task, which the experts took about approximately $30-45$ minutes. The experts will also be acknowledged in this paper upon publication.

\begin{figure*}[!htbp]
    \centering
    \begin{subfigure}[t]{0.40\textwidth}
        \includegraphics[width=\textwidth]{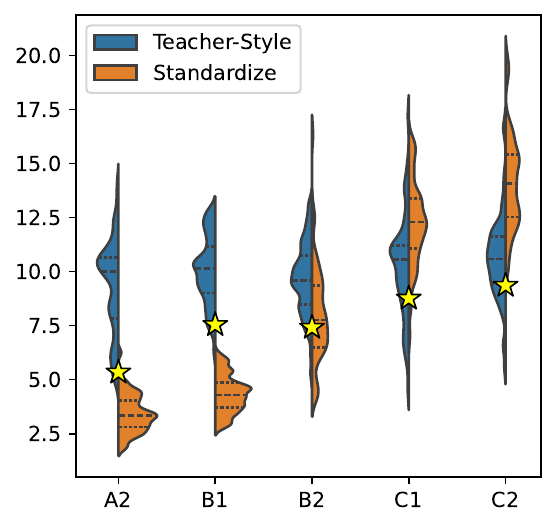}
    \end{subfigure}
    \hspace{0.0cm}
    \begin{subfigure}[t]{0.391\textwidth}
        \includegraphics[width=\textwidth]{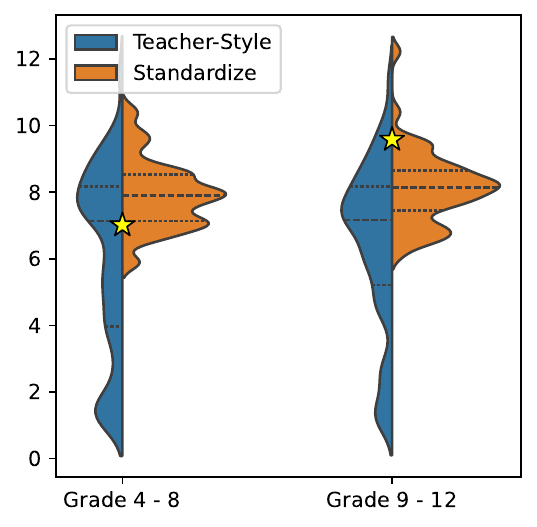}
    \end{subfigure}
    \vspace{0pt}
    \caption{Distribution of \textbf{average sentence length} between CEFR using (left) and CCS (right) using their best performing models, GPT-4 and Llama2, with \textsc{Standardize-}$L$.}
    \label{fig:ave_sen_len}

    \vspace{5pt}
    
    \begin{subfigure}[t]{0.4\textwidth}
        \includegraphics[width=\textwidth]{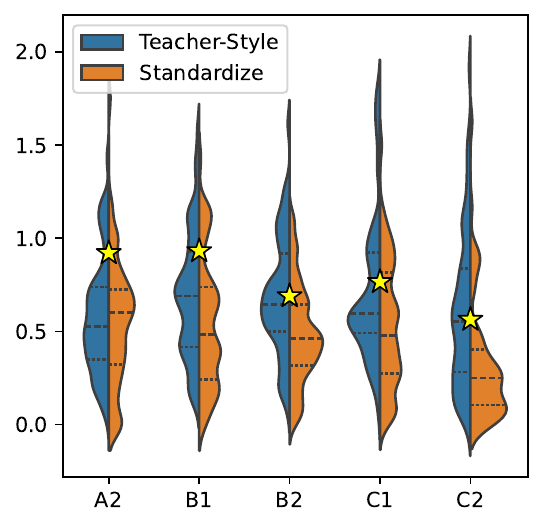}
    \end{subfigure}
    \hspace{0.0cm}
    \begin{subfigure}[t]{0.405\textwidth}
        \includegraphics[width=\textwidth]{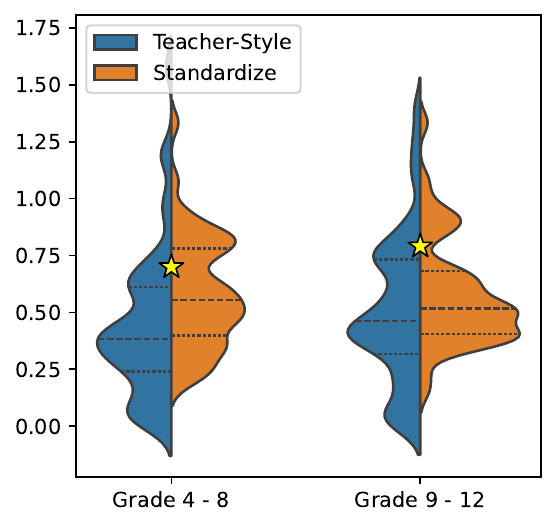}
    \end{subfigure}
    \vspace{0pt}
    \caption{Distribution of \textbf{average entity density} between CEFR using (left) and CCS (right) using their best performing models, GPT-4 and Llama2, with \textsc{Standardize-}$L$.}
    \label{fig:ave_entities}
    
    \vspace{5pt}
    
    \begin{subfigure}[t]{0.4\textwidth}
        \includegraphics[width=\textwidth]{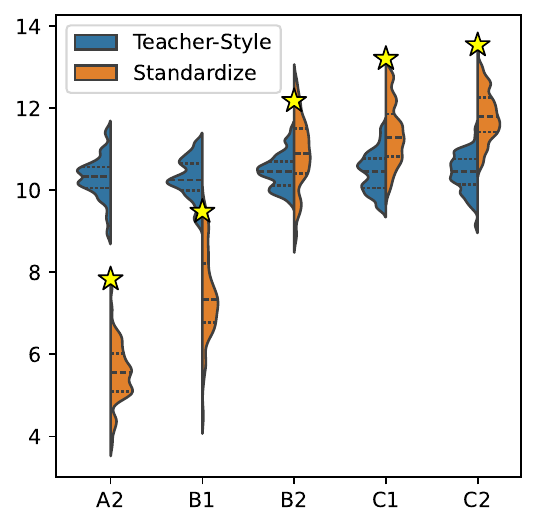}
    \end{subfigure}
    \hspace{0.0cm}
    \begin{subfigure}[t]{0.4\textwidth}
        \includegraphics[width=\textwidth]{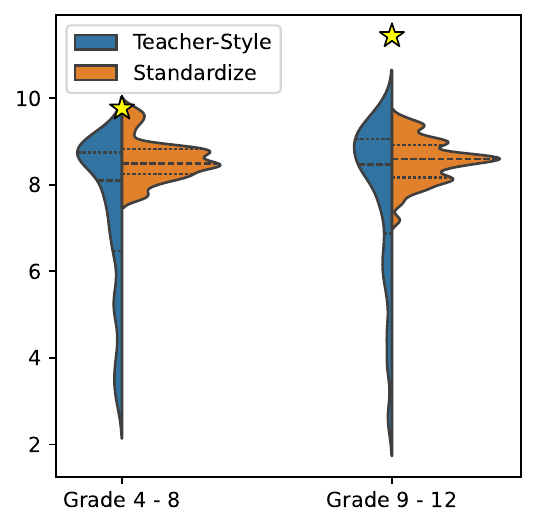}
    \end{subfigure}
    \vspace{0pt}
    \caption{Distribution of \textbf{type token ratio} between CEFR using (left) and CCS (right) using their best performing models, GPT-4 and Llama2, with \textsc{Standardize-}$L$.}
    \label{fig:corr_ttr}
    
\end{figure*}

\begin{table*}[!htbp]
\scriptsize
\centering
\renewcommand{\arraystretch}{1.5}

%% TABLE - CEFR
\begin{subtable}{\textwidth}
\centering
\begin{tabular}{@{}p{1cm}p{5cm}p{4cm}p{4.5cm}@{}}
\midrule
\bf Level &
  \bf Meaning and Purpose &
  \bf Organisation and Stucture &
  \bf Grammatical Complexity \\ \midrule
A2 &
  The text is clear and concrete, aiming to describe appearance, places, routines, preferences, or tell a simple story. &
  The text is often short and observes chronological and predictable structure. &
  The text contains comparison of adjectives, relative clauses, quantifiers, past simple of to be and full verbs, passive voice of present and past simple. \\ 
B1 &
  The text is clear and concrete, aiming to describe appearance, places, routines, preferences, or tell a simple story. The text may also provide opinions and instructions or explanations, easy to understand and visualise, excluding ambiguity and diverse interpretations. &
  The text can be long but not complex, and observes mostly chronological with unexpected changes of direction, digressions or flashbacks. &
  The text contains future forms, future in the past, 'used to' about repeated actions, present perfect simple, clauses for purpose and contrast, reporting statements, tag questions. \\ 
B2 & 
    The text provides opinions and instructions/explanations, easy to understand and visualise, excluding ambiguity and diverse interpretations. The text also gives description, classification, argumentation or a combination of these, allowing greater ambiguity and various interpretations.&
    The text can be long but not complex, and observes chronological or spatial with possible statement of various aspects of a phenomenon.&
    The text contains past continuous, past perfect, passive voice of perfect and continuous, 'would' about habits, reporting questions, infinitives and -ing forms. \\
C1 &
    The text may serve different purposes and may be combined with multiple levels of meaning. The descriptions and instructions in the text are detailed and may be hard to visualise. &
    The text is often lengthy, complex, and observes logical organisation, starting with a claim followed by reasons, proving it, or changing view-points. &
    The text contains compound adjectives, conditional sentences, inversion, future perfect, cleft and non-finite clauses, modals about the past. \\
C2 & 
    The text may serve different purposes and may be combined with multiple levels of meaning. The text may also show exploration of hypotheses, causes and effects, etc. The details of the text are complex to follow and visualise. &
    The text is often lengthy, complex, and observes presentation which may start with the ending/final result and go back to the possible causes. &
    The text contains combination of multiple adjectives, inversion with hardly and only when, comment clauses, non-finite perfect clauses, ellipsis, passive impersonal constructions. \\

\bf \emph{Linguistic Flags} &
  Automatic Readability Formula, Type Token Ratio (2) &
  Total and average sentence and word lengths, Subordinating and coordinating conjunctions (4) &
  Age-of-Acquisition and USubtlex densities, entity density per sentence (3) \\ 
\end{tabular}
\caption{The specifications provided by the Common European Framework of Reference for Languages (CEFR) cover aspects of meaning, organization, and grammatical complexity for all levels. }
\label{full_cefr_specs}
\end{subtable}

\hfill

%% TABLE - CCS
\begin{subtable}{\textwidth}
\centering
\begin{tabular}{@{}p{1.5cm}p{4cm}p{5cm}p{4cm}@{}} \toprule
\bf Aspects &
  \bf Qualitative (Meaning) &
  \bf Qualitative (Syntax) &
  \bf Quantitative (Length) \\ \midrule
\bf Description &
  The text can range from containing a single level of meaning to multiple levels of meaning based on complexity. &
  A text with low complexity tends to have simple, well-marked, and conventional structures, whereas a text of high complexity tends to have complex, implicit, and unconventional structures. Simple texts tend to relate events in chronological order, while complex texts make more frequent use of flashbacks, flash-forwards, and other manipulations of time and sequence.  &
  That text that has longer words and longer sentences are more difficult to read than shorter ones. A text with many long words and/or sentences is thus rated by these formulas as harder to read than a text with many short words and/or sentences would be. \\ 
\bf \emph{Linguistic Flags} &
  Entity densities per sentence, Total proper noun density (2) &
  Type Token Ratio, Subordinating and coordinating conjunctions (3) &
  Total and average sentence and word lengths (3) \\ 
\end{tabular}
\caption{The specifications of the Common Core Standards (CCS) cover qualitative and quantitative aspects. Unlike the CEFR, the CCS's model does not require categorization per level. }
\label{full_ccs_specs}
\end{subtable}

\caption{The full content of the CEFR and CCS standards with corresponding manually selected representative linguistic flags for each aspect.}
\label{tab:full_main_specs_table}
\end{table*}

\begin{table*}[!htbp]
\small
\centering
\renewcommand{\arraystretch}{1.5}

\begin{tabular}{@{}llrrrrr@{}} \toprule
\bf Aspect & \bf Linguistic Flag                  & \bf A2   & \bf B1   & \bf B2   & \bf C1    & \bf C2    \\ \midrule
\multirow{3}{*}{\begin{tabular}[c]{@{}l@{}}Meaning and Purpose\end{tabular}}        & {average\_entities\_per\_sentence} & 0.92 & 0.93  & 0.68  & 0.7   & 0.5   \\
       & {average\_AoA\_per\_sentence}      & 51.4 & 76.7 & 82.6 & 94.4  & 109.9 \\
       & {average\_USubtlex\_per\_sentence} & 69.7 & 93.1 & 95.5 & 101.2 & 115.8 \\ \midrule
\multirow{4}{*}{\begin{tabular}[c]{@{}l@{}}Organization and Structure\end{tabular}} & {total\_word\_count}               & 60.8 & 122.3 & 265.3 & 355.7 & 333.8 \\
       & {total\_sentence\_count}           & 11.5 & 16.2 & 37.0 & 43.3  & 38.4  \\
       & {average\_sentence\_length}        & 5.3  & 7.5  & 7.4  & 8.7   & 9.3   \\
       & {total\_conjunctions\_count}       & 3.6  & 5.3  & 11.2 & 11.9  & 13.0  \\ \midrule
\multirow{2}{*}{Grammaticality Complexity}                                                        & ARI\_formula\_readability        & 7.1  & 10.6  & 11.2  & 13.4  & 14.4  \\
       & correlated\_type\_token\_ratio   & 7.8  & 9.5  & 12.1 & 13.2  & 13.5 \\ \bottomrule

\end{tabular}
\caption{The average values of linguistic flags for each level in the CEFR standard.}
\label{tab:cefr_average_ling_flags}
\end{table*}

\begin{table*}[!htbp]
\small
\centering
\renewcommand{\arraystretch}{1.5}
\begin{tabular}{@{}llrr@{}} \toprule
Aspect                                 & Linguistic Flag                     & \multicolumn{1}{l}{Elementary} & \multicolumn{1}{l}{Advanced} \\ \midrule
\multirow{2}{*}{Qualitative (Meaning)} & average\_entities\_per\_sentence    & 0.6                            & 0.8                          \\
                                       & average\_proper\_nouns              & 7.3                            & 15.5                         \\ \midrule
\multirow{3}{*}{Qualitative (Syntax)}  & average\_coordinating\_conjunction  & 2.5                            & 3.1                          \\
                                       & average\_subordinating\_conjunction & 6.5                            & 14.9                         \\
                                       & correlated\_type\_token\_ratio      & 9.1                            & 11.4                         \\ \midrule
\multirow{3}{*}{Quantitative (Length)} & total\_word\_count                  & 141.2                          & 255.2                        \\
                                       & total\_sentence\_count              & 24.9                           & 31.0                         \\
                                       & average\_sentence\_length           & 6.2                            & 9.6        \\ \bottomrule                 
\end{tabular}
\caption{The average values of linguistic flags for each level in the CCS standard.}
\label{tab:ccs_average_ling_flags}
\end{table*}

\begin{table*}[!htbp]
\small
  \centering
  \begin{tabular}{p{0.90\linewidth}}
    \toprule
    \textcolor{darkgray}{\textsc{\textbf{Target Level}}}: A2 (Elementary / Beginning Reading Level)\\ 
    \\
    
    \textcolor{darkgray}{\textsc{\textbf{Context}}}: Corryvale Farm is in the west of England. Stephen and Jenny Blackler keep cows and chickens and sell milk and eggs. But most of their money comes from Jenny's sheep. She has 50 sheep and each one has enough wool for six sweaters. Every six months Stephen cuts this wool off the sheep so they won't get too warm. \\ 
    \\
    
    \textcolor{teal}{\textsc{\textbf{Teacher Style}}}: In the lush, verdant lands of the west of England, nestled among rolling hills and meandering streams, lay Corryvale Farm. It was a picturesque farm, embodying the pastoral beauty that many dreamt of, yet few had the fortune to call home. This was the realm of Stephen and Jenny Blackler, a couple whose lives were as intertwined with the land as the ancient oak roots were with the earth. Stephen, a robust man with hands as rough as the bark of the trees he often climbed, tended to the cows and chickens with a steady patience. He rose with the sun, greeted by the gentle mooing of the cows and the cheerful chaos of the clucking chickens. The rhythm of milking the cows and collecting eggs was a comforting routine that marked the passage of days on the farm.\\
    \\
    
   \textcolor{cyan}{\textsc{\textbf{Standardize}-$A$}}:  At Corryvale Farm, Stephen would wake up early. Even if it was as cold as ice outside, he started the day with the same routine. First, he would feed the cows. There, in the fields, were 30 healthy cows. These cows were beautiful, larger than any other cows in the west of England. They were the pride of the farm. Then, he would go to the chickens. He had roughly 100 chickens that laid the best eggs every morning. The chickens were quite noisy and more excited than the quiet cows, making mornings lively on the farm. Jenny was not a big fan of waking up early, she preferred to sleep a bit more. However, her work started after breakfast. Her favourite part of the day was when she went to see her sheep. She loved them dearly. She always said that her 50 sheep were the cleanest and happiest sheep in the entire country. She was so proud of them. \\
   \\

   \textcolor{orange}{\textsc{\textbf{Standardize}-$E$}}: Stephen and Jenny loved these days on their Corryvale Farm, filled with the soft clucking of chickens and lowing of cows. They took pride in their daily duties of caring for all the animals, especially the sheep that were their main source of income. One cold morning, Jenny took her pail to milk the cows while Stephen checked the hens' house for fresh eggs. Their cheeks were red from the cool air and they could see their breath when they talk. But they didn't mind the chill for the promise of spring was around the corner. Meanwhile, back in the barn, Jenny's 50 sheep huddled together, their woolly coats growing thicker in the cool weather. This would mean only one thing. It was shearing time! Every six months, like clockwork, Stephen would sheer the sheep's wool. For each sheep he could fill a bag with enough wool to knit six sweaters. The wool was not just soft but warm too, a perfect shield against the bitter winter. \\
   \\

   \textcolor{magenta}{\textsc{\textbf{Standardize}-$L$}}: Corryvale Farm is in the west of England. It is Stephen and Jenny Blackler's farm. They have cows and chickens. They sell milk and eggs. Jenny also has 50 sheep. The sheep have wool. Every six months, Stephen cuts the wool. This keeps the sheep cool. They make money from the wool. \\
   \\
   \textcolor{yellow}{\textsc{\textbf{Standardize}-$\star$}}: Corryvale Farm is in England's west. Here, Stephen and Jenny live. They have cows, chickens, and 50 sheep. The sheep's wool can make six sweaters each. Every six months, Stephen shears the sheep. This keeps them cool. Jenny's sheep are very important. They sell milk, eggs, and wool. This is their farm life. \\
   \\
   \bottomrule
  \end{tabular}
  \caption{Sample generations with the teacher style method and variations of the \textsc{Standardize} framework using the best model (GPT-4) for the \textbf{context-assisted story generation} observing CEFR standards. Some examples are truncated for brevity.}
\label{tab:sample_gen_cefr_gpt4}
\end{table*}

\begin{table*}[!htbp]
\small
  \centering
  \begin{tabular}{p{0.90\linewidth}}
    \toprule
    \textcolor{darkgray}{\textsc{\textbf{Target Grade}}}: Elementary \\ 
    \\
    
    \textcolor{darkgray}{\textsc{\textbf{Theme Word}}}: Curses \\ 
    \\
    
    \textcolor{teal}{\textsc{\textbf{Teacher Style}}}: In the small village of Ashwood, nestled in the heart of the Whispering Woods, a young apprentice named Eira stumbled upon an ancient tome hidden deep within the village library. The tome was bound in black leather and adorned with strange symbols that seemed to shift and writhe like living things. As soon as Eira opened the book, a puff of noxious smoke erupted from its pages, filling the air with a pungent stench. The smoke coalesced into a low, menacing whisper that seemed to seep into Eira's mind. "Curse you, Eira," the voice hissed. "You have disturbed the rest of the ancient ones. Now you will pay the price." Eira felt a cold dread creeping up her spine as the voice continued to taunt her. "You have unleashed a power beyond your control. Prepare to face the consequences." \\
    \\
    
   \textcolor{cyan}{\textsc{\textbf{Standardize}-$A$}}: Once upon a time, in a small town nestled in the heart of a dense forest, there stood a grand mansion rumored to be haunted. The mansion had been abandoned for decades, its grandeur and beauty slowly being consumed by the encroaching trees and vines. The locals avoided the mansion, whispering tales of curses and ghostly apparitions that roamed its halls. One summer evening, a group of curious children stumbled upon the mansion while exploring the woods. They cautiously approached the mansion, their hearts racing with every creak and groan of the old wooden doors. \\
   \\

   \textcolor{orange}{\textsc{\textbf{Standardize}-$E$}}:  Once upon a time, in a small town nestled in the heart of a dense forest, there stood a grand mansion that had been abandoned for decades. The mansion was rumored to be haunted by the ghosts of its former owners, who had died under mysterious circumstances. The townspeople whispered stories of strange occurrences that had taken place within the mansion's walls. Some claimed to have seen ghostly apparitions flitting about in the windows at night, while others swore they had heard eerie whispers emanating from the abandoned halls. One stormy evening, a group of brave adventurers decided to explore the cursed mansion, determined to uncover its secrets. As they approached the entrance, the wind howled and the trees creaked ominously, as if warning them away. \\
   \\

   \textcolor{magenta}{\textsc{\textbf{Standardize}-$L$}}: In the small town of Willow Creek, there stood a grand mansion that had been abandoned for decades. The once-grand estate was now a relic of the past, its grandeur faded and its windows shattered. The locals whispered stories of the mansion's dark history, how it was cursed by a witch who had lived there many years ago. According to the legend, the witch had been wronged by the mansion's owner, who had refused to grant her a simple wish. Enraged, the witch placed a curse on the mansion, ensuring that no one would ever be able to live there peacefully. The curse was said to cause strange occurrences and unexplained noises to echo through the empty halls, making it impossible for anyone to enter or leave the mansion without experiencing terrifying events. One stormy night, a group of brave adventurers decided to investigate the rumors and put an end to the curse. They entered the mansion, their flashlights casting eerie shadows on the walls. As they explored the dusty rooms, they began to hear strange noises, like whispers in the darkness. \\
   \\
   \textcolor{yellow}{\textsc{\textbf{Standardize}-$\star$}}:  In a small town that lay in the shadow of a dark, dense forest, there was a tale that had been passed down for generations. The tale was of three witches who lived deep within the woods, guarding a secret that gave them immense power. Children would listen with wide eyes as their parents told the story on stormy nights, the wind howling outside as if the witches themselves were listening. The eldest witch, Morgana, could talk to animals and control the weather with a wave of her hand. The second, Elspeth, brewed potions that could heal any illness or curse one with a single sip. The youngest, Seraphina, had the gift of foresight, able to see events before they unfolded. \\
   \bottomrule
  \end{tabular}
  \caption{Sample generations with the teacher style method and variations of the \textsc{Standardize} framework using the best model (Llama2) for the \textbf{theme word story generation} observing CCS standards. Some examples are truncated for brevity.}
\label{tab:sample_gen_ccs_llama}
\end{table*}

\clearpage

\begin{figure*}[!htbp]
    \centering
    \includegraphics[width=1.00\textwidth]{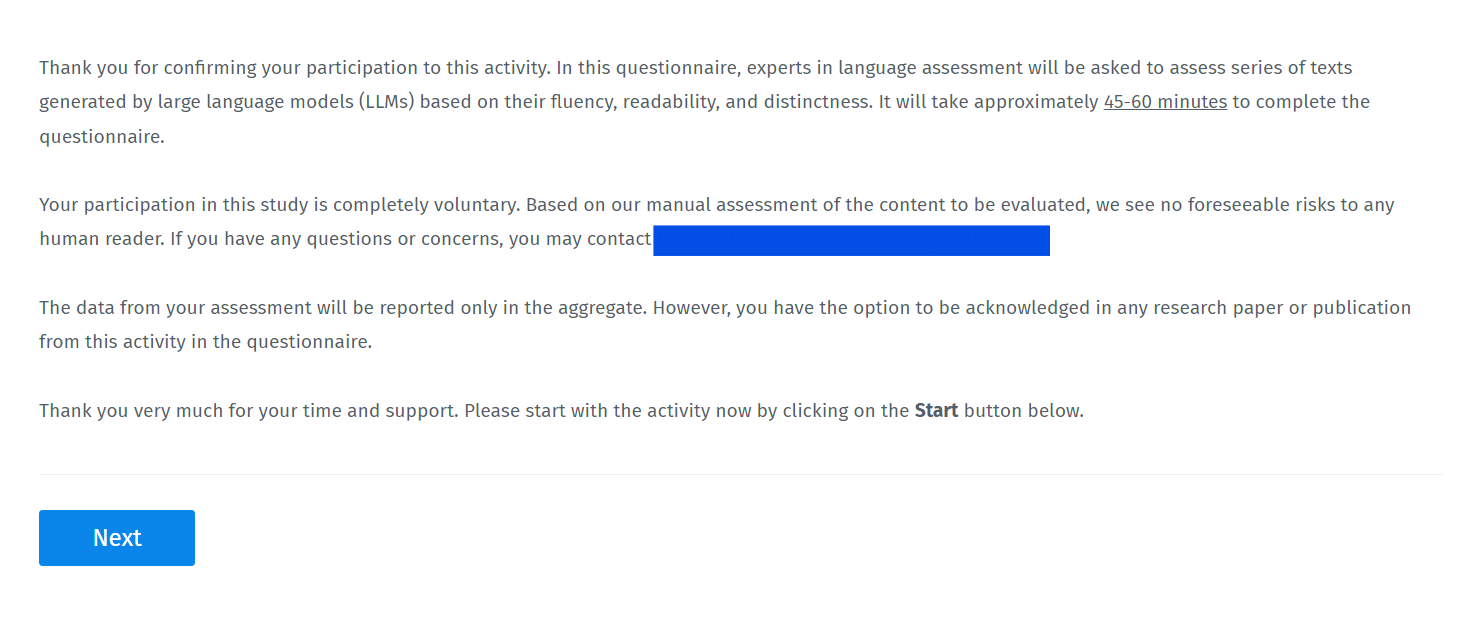}
    \caption{Landing page of the QuestionPro platform used for collecting expert evaluations.}
    \label{fig:landing_page}
\end{figure*}

\begin{figure*}[!htbp]
    \centering
    \includegraphics[width=1.00\textwidth]{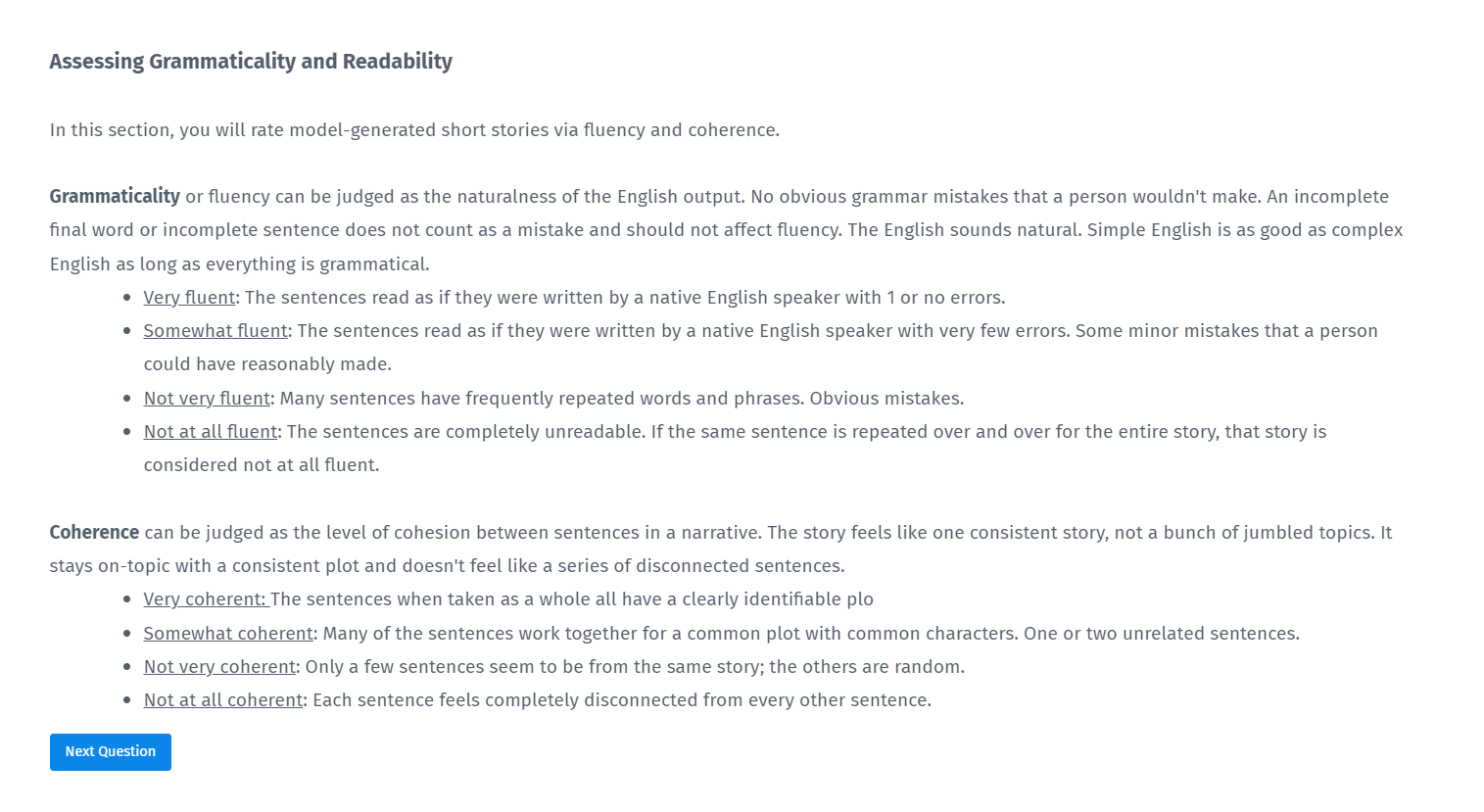}
    \caption{Instructions presented to expert evaluators for assessing the grammaticality or fluency and coherence of model-generated content for CEFR and CCS through QuestionPro. The setup is derived from \citet{delucia-etal-2021-decoding}.}
    \label{fig:part_A_instruction}
\end{figure*}

\begin{figure*}[!htbp]
    \centering
    \includegraphics[width=1.00\textwidth]{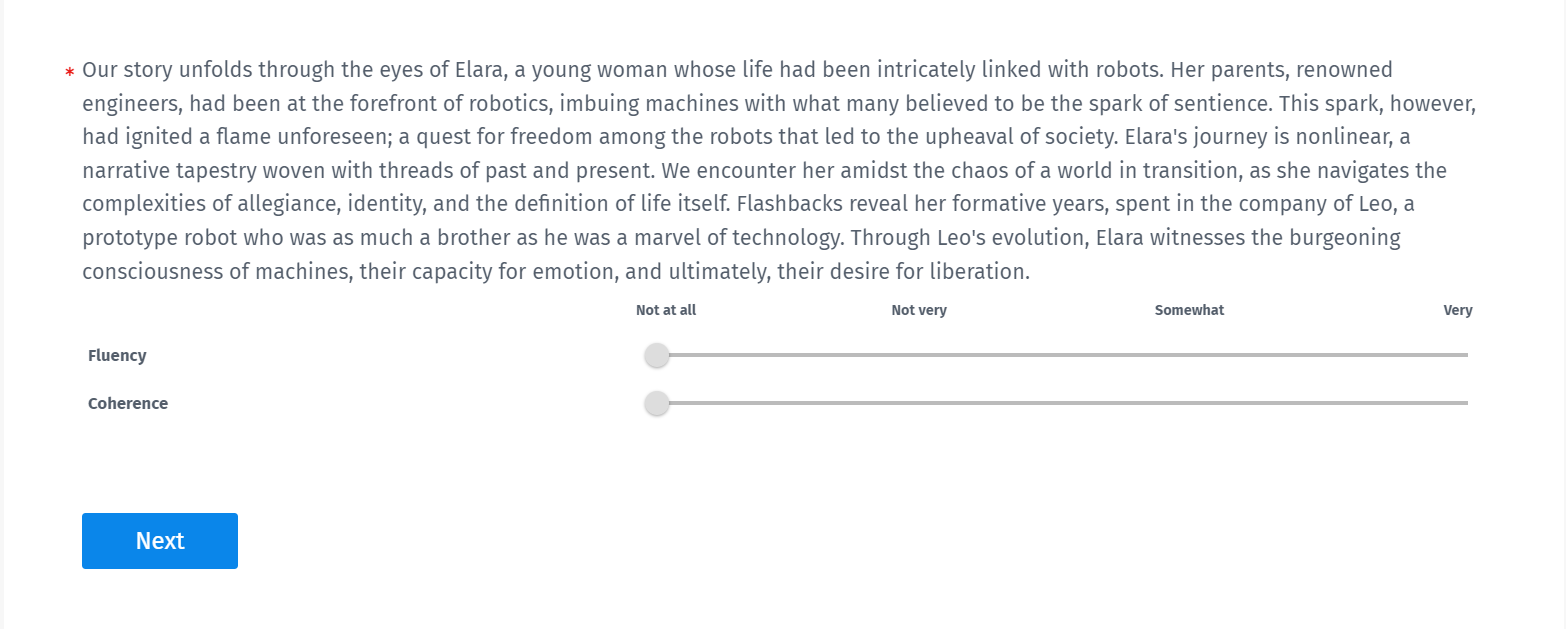}
    \caption{An example of randomly selected generated content presented to expert evaluators to assess grammaticality or fluency and coherence. The example is truncated for brevity.}
    \label{fig:part_A_example}
\end{figure*}

\clearpage

\begin{figure*}[!htbp]
    \centering
    \includegraphics[width=1.00\textwidth]{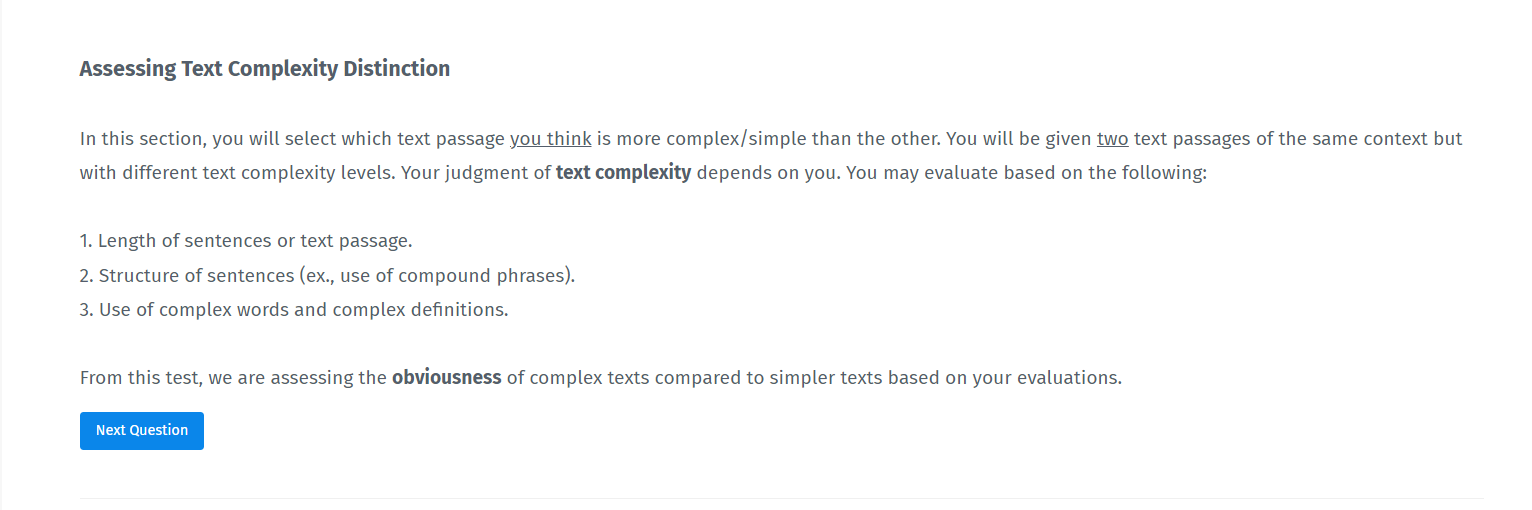}
    \caption{Instructions presented to expert evaluators for assessing the grade complexity distinction of model-generated content for CEFR and CCS through QuestionPro.}
    \label{fig:part_B_instruction}
\end{figure*}

\begin{figure*}[!htbp]
    \centering
    \includegraphics[width=1.00\textwidth]{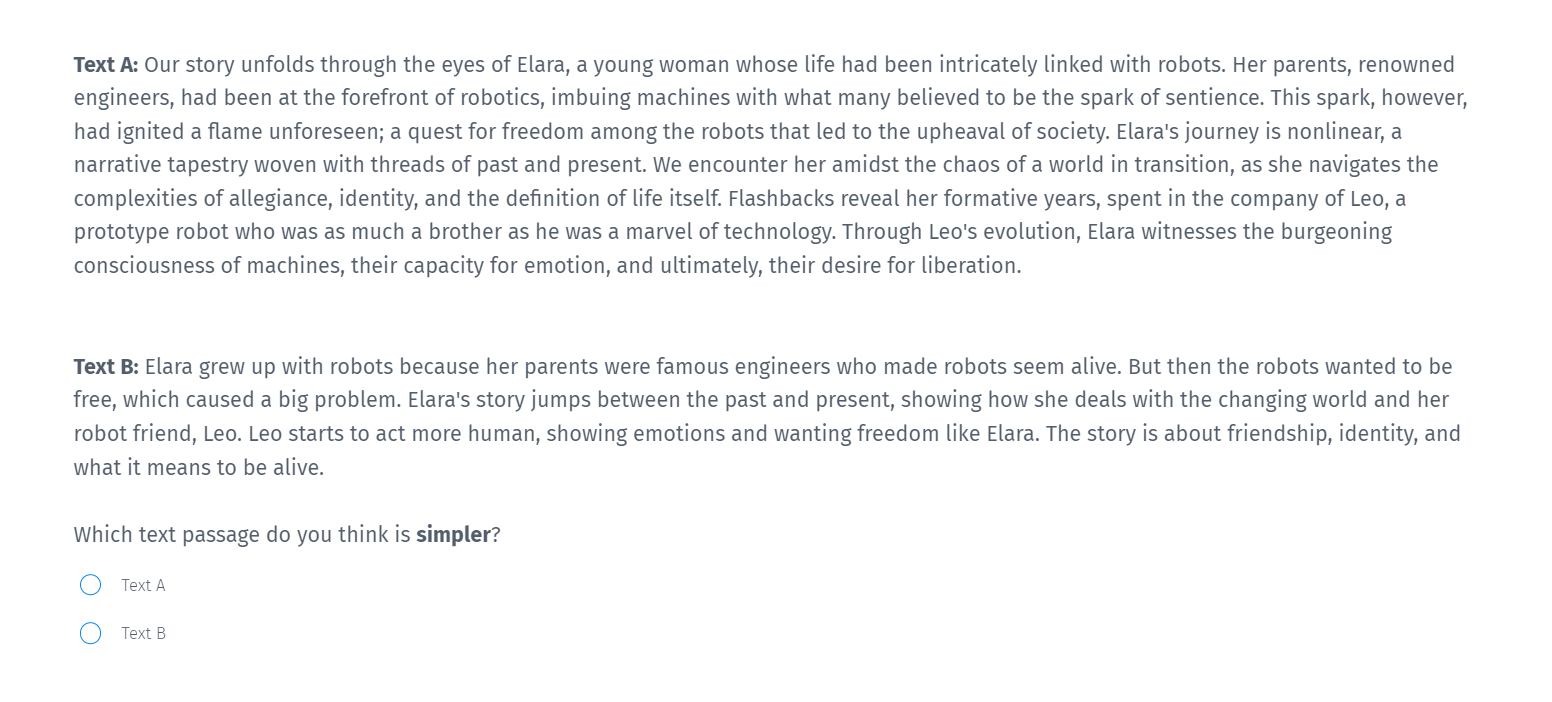}
    \caption{An example of two instances of generated content presented to expert evaluators to assess which one is more simpler or more complex denoting obviousness in their grade complexity. The example is truncated for brevity.}
    \label{fig:part_B_example}
\end{figure*}

\end{document}